\documentclass{article}

\usepackage{etoolbox}
\providetoggle{workingversion}
\settoggle{workingversion}{false}
\providetoggle{blind}
\settoggle{blind}{false}

\newcommand{\ourtitle}{Participatory Personalization in Classification}

\providetoggle{arxiv}
\providetoggle{neurips}
\providetoggle{icml}
\settoggle{arxiv}{false}
\settoggle{neurips}{true}
\settoggle{icml}{false}

\iftoggle{workingversion}{
\usepackage{todonotes}
\let\oldsection\section
\renewcommand\section{\clearpage\oldsection}
}{
\usepackage[disable]{todonotes}
}
\presetkeys{todonotes}{inline, size=\small, color=green!30}{}

\iftoggle{arxiv}{
    \usepackage{ext/jmlr2e_preprint}
    \usepackage[font=small,labelfont=bf]{caption}
    \usepackage{algorithm,algpseudocode}
    \jmlrheading{1}{2023}{}{}{}{Hailey Joren, Chirag Nagpal, Katherine Heller, and Berk Ustun}
    \ShortHeadings{\ourtitle}{Joren et al.}
    \firstpageno{1}
}{}

\iftoggle{neurips}{
    \usepackage[nonatbib,final]{ext/neurips_2023}
    \usepackage[square,comma,sort,numbers]{natbib}
    \usepackage{authblk}
    \title{\ourtitle{}}
    \author[1]{Hailey Joren}
    \author[2]{Chirag Nagpal}
    \author[2]{Katherine Heller}
    \author[1]{Berk Ustun}
    \affil[1]{UC San Diego}\affil[2]{Google}
    \usepackage[font=footnotesize,labelfont=bf]{caption}
    \usepackage{algorithm,algpseudocode}
}{}

\usepackage[utf8]{inputenc} 
\usepackage[T1]{fontenc}    
\usepackage{microtype}      

\usepackage{environ,comment}
\usepackage{amsfonts,bbm,bm,courier}
\usepackage{amsmath,amsthm,amssymb,nicefrac}

\usepackage{hyperref}    
\usepackage{graphicx,placeins,xcolor,caption}
\usepackage{array,booktabs,tabularx,multirow,colortbl,hhline}
\usepackage{adjustbox}
\usepackage{enumitem}

\definecolor{myyellow}{RGB}{250,165,32}
\definecolor{mydarkgreen}{RGB}{0,180,0}
\definecolor{mydarkred}{RGB}{100,0,0}

\usepackage{wasysym}
\usepackage{makecell}
\usepackage{tikz,pgfplots}
\usepackage[linguistics,edges]{forest}
\usepackage{longtable}
\usepackage{wrapfig}
\usepackage{float}
\usepackage{pdflscape}
\usepackage{threeparttable}
\usepackage{threeparttablex}
\usepackage[normalem]{ulem}
\usepackage{flafter}
\pgfplotsset{compat=1.17}

\hypersetup{colorlinks=true, urlcolor=blue, linkcolor=blue, citecolor=blue, linkbordercolor=blue, pdfborderstyle={/S/U/W 1}}

\usepackage[capitalise]{cleveref}

\newcolumntype{H}{>{\setbox0=\hbox\bgroup}c<{\egroup}@{}}

\newcommand{\textds}[1]{{\footnotesize{\texttt{#1}}}} 
\newcommand{\textgp}[1]{{\footnotesize{\texttt{#1}}}} 
\newcommand{\textmn}[1]{\textsf{\small{#1}}} 

\newcommand{\textheader}[1]{{\bfseries{#1}}}
\newcommand{\cell}[2]{\setlength{\tabcolsep}{0pt}\begin{tabular}{#1}#2 \end{tabular}}

\setitemize{leftmargin=0.5em,topsep=0.1em,parsep=0.5pt,label=\raisebox{0.25ex}{\tiny$\bullet$}}
\setlist[enumerate]{leftmargin=*, label= {\arabic*.}, itemsep=0.5em}

\newtheorem{definition}{Definition}

\algnewcommand\algorithmicinput{\textbf{Input}}
\renewcommand\algorithmicensure{\textbf{Output}}
\algnewcommand\algorithmicinitialize{\textbf{Initialize}}
\algnewcommand\algorithmicbigstep{\textbf{Step}}
\algnewcommand\INPUT{\item[\algorithmicinput]}
\algnewcommand\INITIALIZE{\item[\algorithmicinitialize]}
\algnewcommand{\STEP}[1]{\item[\algorithmicbigstep]{\textbf{#1}}}
\algnewcommand{\InputExplanation}[2][.6\linewidth]{\leavevmode\hfill\makebox[#1][r]{~{\footnotesize{#2}}}}
\algnewcommand{\InitializationExplanation}[2][.6\linewidth]{\leavevmode\hfill\makebox[#1][r]{~{\footnotesize{#2}}}}
\algnewcommand{\alginput}[2]{\Statex{Input:~#1}\Comment{#2}}
\algnewcommand{\StateComment}[2]{\State{#1}\InputExplanation{#2}}
\algnewcommand{\alginitialize}[2]{\Statex{#1}\InitializationExplanation{#2}}
\algrenewcommand\algorithmiccomment[2][]{#1\hfill\textit{\scriptsize{#2}}}

\usepackage{listofitems}
\usepackage[toc,page,header]{appendix}
\usepackage{minitoc}
\makeatletter
\patchcmd{\hyper@makecurrent}{%
    \ifx\Hy@param\Hy@chapterstring
        \let\Hy@param\Hy@chapapp
    \fi
}{%
    \iftoggle{inappendix}{
        \@checkappendixparam{chapter}%
        \@checkappendixparam{section}%
        \@checkappendixparam{subsection}%
        \@checkappendixparam{subsubsection}%
        \@checkappendixparam{paragraph}%
        \@checkappendixparam{subparagraph}%
    }{}%
}{}{\errmessage{failed to patch}}

\newcommand*{\@checkappendixparam}[1]{%
    \def\@checkappendixparamtmp{#1}%
    \ifx\Hy@param\@checkappendixparamtmp
        \let\Hy@param\Hy@appendixstring
    \fi
}
\makeatletter

\makeatletter
\newcommand*{\centerfloat}{%
  \parindent \z@
  \leftskip \z@ \@plus 1fil \@minus \textwidth
  \rightskip\leftskip
  \parfillskip \z@skip}
\makeatother

\newtoggle{inappendix}
\togglefalse{inappendix}
\apptocmd{\appendix}{\toggletrue{inappendix}}{}{\errmessage{failed to patch}}
\apptocmd{\subappendices}{\toggletrue{inappendix}}{}{\errmessage{failed to patch}}

\definecolor{best}{HTML}{BAFFCD}
\definecolor{bad}{HTML}{FFC8BA}

\newcommand{\badvalue}[1]{{\color{red}{\textbf{#1}}}} 

\definecolor{fgood}{HTML}{BAFFCD}
\definecolor{fbad}{HTML}{FFC8BA}
\definecolor{funk}{HTML}{FFFFE0}
\definecolor{darkfunk}{HTML}{DDDD00}

\newcommand{\ftblgap}{\quad}
\newcommand{\ftblheader}[3]{\multicolumn{#1}{#2}{\cell{#2}{#3}}}

\newcommand\clearrow{\global\let\rowmac\relax}

\newcommand{\R}{\mathbb{R}}
\newcommand{\E}[0]{\mathbb{E}}

\DeclareMathOperator*{\argmax}{argmax}
\DeclareMathOperator*{\argmin}{argmin}

\newcommand{\data}[1]{\mathcal{D}_{#1}}

\newcommand{\xb}{\bm{x}}
\newcommand{\X}{\mathcal{X}}
\newcommand{\Y}{\mathcal{Y}}

\newcommand{\txplus}[0]{+}
\newcommand{\txminus}[0]{-}

\newcommand{\nplus}[1]{n^{\txplus{}}_{#1}}
\newcommand{\nminus}[1]{n^{\txminus{}}_{#1}}
\newcommand{\n}[1]{n_{#1}}

\newcommand{\gb}{\bm{g}}
\newcommand{\G}{\mathcal{G}}

\newcommand{\clf}[0]{h_0}
\newcommand{\plf}[1]{h_{#1}}

\newcommand{\Hset}[0]{\mathcal{H}}

\newcommand{\truerisk}[2]{{R_{#1}(#2)}}
\newcommand{\emprisk}[2]{{\hat{R}_{#1}(#2)}}

\newcommand{\empgap}[3]{\hat{\Delta}_{#1}({#2},{#3})}
\newcommand{\truegap}[3]{\Delta_{#1}({#2}, {#3})}

\newcommand{\dnr}{\varnothing}
\newcommand{\Dnr}[0]{\bm{\dnr}}
\newcommand{\RG}{\mathcal{R}}
\newcommand{\rg}{\bm{r}}

\newcommand{\candidatepool}{\mathcal{M}}

\newcommand{\trees}[1]{\mathbbm{T}}

\newcommand{\supergroups}[1]{\textsf{supergroups}(\gb)}

\newcommand{\metricsguide}[0]{}

\newcommand{\nodestats}[5]{}

\forestset{
 prunednode/.style={text=gray, draw, gray, fill=white, font=\Large, dashed, outer sep=.3em, inner sep=0.8em},
 genericnode/.style={text=black, draw, fill=lightgray, font=\Large, inner sep=0.8em, outer sep=0.2em},
 personnode/.style n args={2}{text=black, draw, fill=cyan!#1, font=\Large, outer sep=1.2em, inner sep=0.8em},
 groupattr/.style={text=red,font=\ttfamily\Large, s sep=2mm, l sep=15mm},
 leaflabel/.style={edge={white}, edge label={node[text=black, above,font=\ttfamily\Large,fill=white,inner sep=2pt]{}}},
 leaflabelpruned/.style={edge={white}, edge label={node[text=gray, above,font=\ttfamily\Large,fill=white,inner sep=2pt]{#1}}},
 grouplabel/.style={edge label={node[above,font=\ttfamily\Large,fill=white,inner sep=2pt]{#1}}},
 grouplabelpruned/.style={edge={dashed, gray}, edge label={node[above,font=\ttfamily\Large,fill=white,inner sep=2pt, text=gray]{#1}}}
}

\begin{document}
\doparttoc\faketableofcontents

\iftoggle{neurips}{\maketitle}{}

\iftoggle{arxiv}{
\title{\ourtitle{}}
\author{%
\name Hailey Joren \email hjoren@ucsd.edu \\ \addr University of California, San Diego \AND
\name Chirag Nagpal \email chiragn@cs.cmu.edu \\ \addr Google \AND
\name Katherine Heller \email kheller@google.com \\ \addr Google \AND
\name Berk Ustun \email berk@ucsd.edu \\ \addr University of California, San Diego%
}
\maketitle
}

\iftoggle{icml}{
\iftoggle{workingversion}{
\icmltitlerunning{\ourtitle\hfill\thepage}
}{}
\twocolumn[
\icmltitle{\ourtitle{}}
\icmlsetsymbol{equal}{*}
\begin{icmlauthorlist}
\icmlauthor{Hailey Joren}{ucsd}
\icmlauthor{Chirag Nagpal}{cmu}
\icmlauthor{Katherine Heller}{google}
\icmlauthor{Berk Ustun}{ucsd}
\end{icmlauthorlist}
\icmlaffiliation{ucsd}{UCSD}
\icmlaffiliation{cmu}{Carnegie Mellon University}
\icmlaffiliation{google}{Google}
\icmlcorrespondingauthor{Hailey Joren}{hjoren@ucsd.edu}
\icmlkeywords{Informed Consent, Personalization, Participation, Data Privacy, Algorithmic Fairness, Healthcare, Clinical Decision Support}
\vskip 0.3in
]
\printAffiliationsAndNotice{}
}

\begin{abstract}
Machine learning models are often personalized with information that is protected, sensitive, self-reported, or costly to acquire. These models use information about people but do not facilitate nor inform their \emph{consent}. Individuals cannot opt out of reporting personal information to a model, nor tell if they benefit from personalization in the first place. We introduce a family of classification models, called \emph{participatory systems}, that let individuals opt into personalization at prediction time. We present a model-agnostic algorithm to learn participatory systems for personalization with categorical group attributes. We conduct a comprehensive empirical study of participatory systems in clinical prediction tasks, benchmarking them with common approaches for personalization and imputation. Our results demonstrate that participatory systems can facilitate and inform consent while improving performance and data use across all groups who report personal data.
\end{abstract}

\iftoggle{arxiv}{
\vspace{0.25em}
\begin{keywords}
Informed Consent, Personalization, Participation, Data Privacy, Algorithmic Fairness, Healthcare, Clinical Decision Support
\end{keywords}
}

\section{Introduction}
\label{Sec::Introduction}

Machine learning models routinely assign predictions to \emph{people} -- be it to screen a patient for a mental illness~\cite{kessler2005world}, their risk of mortality in an ICU~\cite{pollard2018eicu}, or their likelihood of responding to treatment~\cite{abajian2018predicting}. Many models in such applications are designed to target heterogeneous subpopulations using features that explicitly encode personal information. Typically, models are \emph{personalized} with categorical attributes that define groups~\citep[i.e., ``categorization'' as per][]{fan2006personalization}. In medicine, for example, clinical prediction models use \emph{group attributes} that are \emph{protected} (e.g., \textgp{sex} in the \href{https://www.mdcalc.com/calc/801/cha2ds2-vasc-score-atrial-fibrillation-stroke-risk}{CHA\textsubscript{2}DS\textsubscript{2} Score for Stroke Risk}), \emph{sensitive} (e.g., \textgp{HIV} status in the \href{https://www.mdcalc.com/calc/10346/veterans-health-administration-covid-19-vaco-index-covid-19-mortality}{VA COVID-19 Mortality Score}), \emph{self-reported} (e.g., \textgp{alcohol\_use} in the
\href{https://www.mdcalc.com/calc/807/has-bled-score-major-bleeding-risk}{HAS-BLED Score for Major Bleeding Risk}), or \emph{costly} to acquire (e.g., \textgp{leukocytosis} in the \href{https://www.mdcalc.com/calc/617/alvarado-score-acute-appendicitis}{Alvarado Appendicitis Score}).

\begin{figure*}[!t]
\centering
\resizebox{1.0\linewidth}{!}{
\begin{tabular}{c>{\;}r>{\;\;}r>{\ftblgap{}}r>{\;\;}r>{\ftblgap{}}r>{\;\;}r}
    \ftblheader{7}{c}{}\\[1.5em] 
    \ftblheader{1}{c}{Group} & \ftblheader{2}{c}{Data} & \ftblheader{2}{c}{Personalized} & \ftblheader{2}{c}{Generic}\\
    \cmidrule(lr){1-1}\cmidrule(lr){2-3}\cmidrule(lr){4-5}\cmidrule(lr){6-7}
    \ftblheader{1}{c}{$\gb{}$} &  $\nplus{\gb}$ & $\nminus{\gb}$  & $\plf{}$ & $R_{\gb}(\plf{})$ & $\clf{}$ & $R_{\gb}(\clf)$\\
    \cmidrule(lr){1-1}\cmidrule(lr){2-3}\cmidrule(lr){4-5}\cmidrule(lr){6-7} 
     \textgp{female}, \textgp{old} &  $0$ & $24$ &  $+$ & 24 & $-$ & 0 \\       
     \textgp{female}, \textgp{young} &  $25$ & $0$ & $+$ & 0 & $-$ & 25 \\
     \textgp{male}, \textgp{old} &  $25$ & $0$ &   $+$ & 0 & $-$ & 25 \\
    \textgp{male}, \textgp{young} &   $0$ & $27$ &  $-$ & 0 & $-$ & 0 \\
     \cmidrule(lr){1-1}\cmidrule(lr){2-3}\cmidrule(lr){4-5}\cmidrule(lr){6-7}
     \hspace{4em}\textbf{Total} &  $50$ & $51$ &  & 24 & & 50
\end{tabular}
\begin{tabular}{ccc}
    \ftblheader{3}{c}{\textsf{Traditional Personalization}}\\
    \ftblheader{3}{c}{\color{gray}\textsf{groups receive predictions from $\plf{}$}}\\[0.5em]
    \ftblheader{1}{c}{Model}  &  \ftblheader{1}{c}{Data Use} & \ftblheader{1}{c}{Gain}\\
    \cmidrule(lr){1-3}
    & $\rg$ & $\Delta R_{\gb}(\plf{}, \clf{})$ \\
    \cmidrule(lr){1-1}  \cmidrule(lr){2-2}  \cmidrule(lr){3-3} 
     $\plf{}$ & \textgp{female}, \textgp{old} &  \cellcolor{fbad} $-24$ \\       
     $\plf{}$ & \textgp{female}, \textgp{young} & \cellcolor{fgood} $25$ \\
     $\plf{}$ & \textgp{male}, \textgp{old}  &  \cellcolor{fgood} $25$ \\
     $\plf{}$ & \textgp{male}, \textgp{young} &  \cellcolor{funk} $0$ \\ 
    \cmidrule(lr){3-3} 
    & &  \cellcolor{fgood} $26$
\end{tabular}
\begin{tabular}{ccc}
    \ftblheader{3}{c}{\textsf{Minimal Participatory System}}\\
    \ftblheader{3}{c}{\color{gray}\textsf{groups opt into predictions from $\plf{}$ or $\clf{}$}}\\[0.5em]
    \ftblheader{1}{c}{Model}  &  \ftblheader{1}{c}{Data Use} & \ftblheader{1}{c}{Gain} \\
    \cmidrule(lr){1-3}
    & $\rg$ & $\Delta R_{\gb}(\plf{}, \clf{})$ \\
    \cmidrule(lr){1-1}  \cmidrule(lr){2-2}  \cmidrule(lr){3-3} 
     $\clf{}$ &  \cellcolor{fgood} $\dnr{}$ &   $0$   \\       
     $\plf{}$ &  \textgp{female}, \textgp{young} &   \cellcolor{fgood} $25$   \\  
     $\plf{}$ &  \textgp{female}, \textgp{young} &   \cellcolor{fgood} $25$   \\  
     $\clf{}$ &  \cellcolor{fgood} $\dnr{}$ &  $0$  \\  
    \cmidrule(lr){3-3} 
     &  &   \cellcolor{fgood} $50$  
\end{tabular}}
\caption{Classification task where participation improves accuracy and minimizes data use. 
%
%
%
Under traditional personalization, individuals report group membership to receive personalized predictions from $\plf{}$. As shown, this benefits the population as a whole by improving overall error from 50 to 24 ($\Delta R_{\gb}(\plf{}, \clf{})=$\textbf{{\color{mydarkgreen}$~26$}}). However, it has no effect on $[\textgp{male}, \textgp{young}]$, who receive the same predicitons ($\Delta R_{\gb}(\plf{}, \clf{})=$\textbf{{\color{myyellow}$~0$}}) and a detrimental effect to$[\textgp{female}, \textgp{old}]$, who receive less accurate predictions ($\Delta R_{\gb}(\plf{}, \clf{})=~$\textbf{{\color{red}$-24$}}).
In a minimal participatory system, individuals can \emph{opt in} to personalization to receive predictions from $\plf{}$ or $\clf{}$. Here, individuals in groups $[\textgp{female}, \textgp{old}]$ and $[\textgp{male}, \textgp{young}]$ would opt out of personalization, leading to an overall error of 0 ($\Delta R_{\gb}(\plf{}, \clf{})=~$\textbf{{\color{mydarkgreen}$50$}}) and a reduction in unnecessary data collection (\textbf{{\color{mydarkgreen}$\dnr$}}).}
\label{Fig::MinimalParticipation}
\end{figure*}

Websites that solicit personal data are often required to support \emph{informed consent}: individuals can opt out of providing personal data and understand how it will be used~\citep[see, e.g., personal data guidelines in GDPR, OECD privacy guidelines][]{gdpr,oecd2013}.
Outside of these contexts, personalized models do not provide such functionality: individuals cannot opt out of reporting data used to personalize their predictions nor tell if it would improve their predictions.
Practitioners assume that data available for training will be available at inference time. In practice, this assumption has led to a proliferation of models that use information that individuals may be unwilling or unable to report at prediction time~\citep[see e.g., the \href{https://www.mdcalc.com/calc/10046/denver-hiv-risk-score}{Denver HIV Risk Score}][which asks patients to report \textgp{age}, \textgp{gender}, \textgp{race}, and \textgp{sexual\_practices}]{denverhiv}. In tasks where individuals self-report, they may not voluntarily report information that could improve their predictions or may report incorrect information. 

The broader need to facilitate and inform consent in personalization stems from the fact that it may not improve performance for each group that reports personal data~\citep[][]{suriyakumar2022fairuse}. In practice, a personalized model can perform \emph{worse} or the same as a \emph{generic model} fit without personal information for a group with specific characteristics. Such models violate the implicit promise of personalization as individuals report personal information without receiving a tailored performance gain in return. These instances of \emph{worsenalization} are prevalent, hard to detect, and hard to resolve~\citep[see][]{suriyakumar2022fairuse,paes2022on}. However, they would be resolved if individuals could opt out of personalization and understand its expected gains (see \cref{Fig::MinimalParticipation}).

In this work, we introduce a family of classification models that operationalize informed consent called \emph{participatory systems}. Participatory systems \emph{facilitate consent} by allowing individuals to report personal information at prediction time. Moreover, they \emph{inform consent} by showing how reporting personal information will change their predictions. Models that facilitate consent operate as \emph{markets} in which individuals trade personal information for performance gains. 
This work seeks to develop systems that (i) perform as well as possible when individuals opt-in (to incentivize voluntary reporting) and opt out (to safeguard against abstention); (ii) provide opportunities for individuals to make informed decisions about data provision.
The resulting systems can improve performance and curb data use for all groups who report personal data, maximizing the gains of personal information when it improves performance and limiting data collection when it does not.

%






\paragraph{Related Work}
\newcommand{\inlineheader}[1]{\noindent\emph{#1}.\quad\nolinebreak}

Participatory systems support modern principles of responsible data use such as \emph{informed consent} and \emph{collection limitation} -- i.e., data should be collected with a data subject's consent and restricted to only what is necessary. These principles are articulated in, e.g., OECD privacy guidelines~\citep{oecd2013}, the GDPR~\citep{gdpr}, and the California Consumer Privacy Act~\citep{ccpa}. These principles stem from extensive work on the right to data privacy~\citep[][]{kaminski2019right}.
They are motivated, in part, by research showing that individuals care deeply about their ability to control personal data~\citep{bansal2010impact, anderson2011digitization, auxier2019americans} and differ considerably in their desire or capacity to share it~\citep[see e.g.][]{awad2006personalization,ortlieb2016sensitivity,campbell1980information, chemmanur1993pricing,auer2008near,lundberg2019privacy, arellano2018privacy}. We follow an interpretation of these principles that ties data minimization to system performance, and our work complements efforts to understand the trade-off between personalization and performance~\cite{biega2020operationalizing, shanmugam2022learning}.
%
 


We consider models that are personalized with categorical attributes that encode personal characteristics
\citep[i.e., ``categorization'' rather than ``individualization'' as per][]{fan2006personalization}. 
Modern techniques for learning with categorical attributes~\citep[see e.g.,][]{agresti2018introduction,steyerberg2019clinical} use them to improve performance at a population level -- e.g., by accounting for higher-order interaction effects~\citep{bien2013lasso,lim2015learning,vaughan2020efficient} or recursive partitioning~\citep{elmachtoub2018value,biggs2020model,bertsimas2019optimal,bertsimas2020predictive}. Our proposed methods can be used to achieve these goals in tasks where models use features that are optional or costly to acquire~\citep[see e.g.,][]{schaar_costly,auer2008near,Yu2009ActiveS,tran2023personalized}.


Our work is broadly related to algorithmic fairness in that we seek to improve model performance at a group level. Recent work shows that personalization with group attributes does not uniformly improve performance and can result in less accurate predictions at a group level~\citep[see][]{suriyakumar2022fairuse,paes2022on, ustun2019fairness}. The proposed systems can mitigate this effect by allowing individuals to opt out of these instances of ``\emph{worsenalization}.'' This line of work complements research on preference-based fairness~\citep{zafar2017parity,ustun2019fairness,kim2019preference,viviano2020fair,do2021online}, on ensuring fairness across complex group structures~\citep[]{kearns2018preventing,hebert2018multicalibration,2022biasbounties}, and on the study of privacy across subpopulations~\citep{tran2021differentially, biega2020operationalizing}.

\section{Participatory Systems}
\label{Sec::ProblemStatement}

We consider a classification task where we personalize a model with categorical attributes. We start with a dataset $\{(\xb_i, y_i, \gb_i)\}_{i=1}^n$ where each example consists of a feature vector $\xb_i \in \R^{d}$, a label $y_i \in \Y$, and a vector of $k$ categorical attributes $\gb_i = [g_{i,1},\ldots,g_{i,k}] \in \G_1 \times \ldots \times \G_k = \G$. We refer to $\G$ as \emph{group attributes} and $\gb_i$ as the \emph{group membership} of person $i$. We let $\n{\gb} := |\{i ~| \gb_i = \gb \}|$ denote the size of group $\gb$, and $|\G_{i}|$ denote the number of categories for group attribute $i$.

We use the dataset to fit a model $\plf{}:\X \times \G \to \Y$ by empirical risk minimization with a loss $\ell: \Y\times\Y\to\R_+$.  Given a model $\plf{}$, we denote its true risk and empirical risk $\truerisk{}{\plf{}}$ and $\emprisk{}{\plf{}}$.  
We denote the true risk and empirical risk of $\plf{}$ for group $\gb \in \G$ when they are assigned the personalized predictions for group $\gb' \in \G$ as:
\begin{align*}
\truerisk{\gb}{\plf{}(\cdot, \gb')} := \E\left[\ell\left(\plf{}(\cdot,\rg'),y \right) \mid \G =\gb'\right], \qquad 
\emprisk{\gb}{\plf{}(\cdot,\gb')} := \frac{1}{\n{\gb}} \sum_{i:\rg_i = \gb}\ell\left(\plf{}(\cdot,\rg'),y_i \right).
\end{align*}

We assume that groups prefer more accurate predictions. Given models $\plf{}$ and $\plf{}'$, users in group $\gb{}$ prefer $\plf{}$ when $\truerisk{\gb}{\plf{}} < \truerisk{\gb}{\plf{}'}$. This assumption holds in settings where individuals prefer more accurate predictions -- e.g., when predicting the risk of an illness~\citep[see, e.g.,][]{struck2020assessment,le1993new,struck2017association} or recommending content~\cite{mcauley2022pml}. It does not hold in applications where some individuals prefer inaccurate predictions -- e.g., ``polar'' clinical prediction tasks such as predicting the risk of organ failure for a transplant~\citep[][]{paulus2020predictably}. 


We consider models where individuals consent to personalization by reporting group attributes at prediction time. We let $\dnr{}$ denote an attribute that was not reported, and let $\rg_i = [r_{i,1},\ldots,r_{i,k}] \in \RG \subseteq \G \times \Dnr{}$ denote the \emph{reported group membership} of person $i$. For example, a person with $\gb_i = [\textgp{female},\textgp{HIV}=\textgp{+}]$ would have $\rg_i = [\textgp{female}, \dnr{}]$ if they only report $\textgp{sex}$, and $\rg_i = \Dnr{} := [\dnr, \ldots, \dnr]$ if they opt out of personalization entirely. 

We associate each model with a set of \emph{reporting options} $\RG$. A traditional model that requires each person to report group attributes has $\RG = \G$. A model where each person could report any subset of group attributes has $\RG = \G \times \Dnr{}$. The truthful reporting options of a group $\gb$ include all $\rg \in \dim{\gb} := \gb \times \Dnr$. We use $n_{\rg}$ to denote the number of individuals who could truthfully report $\rg \in \RG{}$. We represent individual decisions to opt into personalization at prediction time through a \emph{reporting interface}, defined below.



\newcommand{\rtree}{T}
\newcommand{\parent}[1]{\textnormal{pa}(#1)}
\renewcommand{\root}[1]{\textnormal{root}(#1)}
\newcommand{\nodes}[1]{\textnormal{nodes}(#1)}
\newcommand{\leaves}[1]{\textnormal{leaves}(#1)}
\newcommand{\psys}[1]{f_{#1}}
\newcommand{\rmodels}{\mathcal{M}}

\begin{definition}\normalfont\label{Def::ReportingInterface}
Given a personalized classification task with group attributes $\G$, a \emph{reporting interface} is a tree $\rtree$ whose $\nodes{\rtree} = \RG \subseteq \G \times \Dnr{}$ represent attributes reported at prediction time. 
The tree is rooted at $\root{\rtree} = \Dnr{}$ and branches as an individual reports group attributes for personalization. Given a node $\rg$, we denote its child $\rg'$ as $\rg = \parent{\rg'}$. Each parent-child pair represents a \emph{reporting decision}, and the tree's height represents the maximum number of reporting decisions. 
\end{definition}

We say that a tree supports \emph{consent} to personalization if users can opt out of personalization entirely $\Dnr \in \textrm{nodes}(\rtree)$. 




\begin{definition}
\normalfont
Given a personalized classification task with group attributes $\G$, a \emph{participatory system} with reporting interface $\rtree$ is a prediction model $\psys{\rtree}: \X \times \RG \to \Y$ that obeys the following properties:
\begin{enumerate}[leftmargin=*, itemsep = 0pt, label={(R\arabic*)}]

    \item \emph{Incentive Compatibility}: Opting into personalization improve expected performance 
    \begin{align*}
    \truerisk{\rg}{\psys{\rtree}({\cdot,\rg})} < \truerisk{\rg}{\psys{\rtree}({\cdot,\rg}')} 
    \textrm{ for all reporting groups }
    \rg, \rg' \textrm{ such that } \rg' = \parent{\rg}
    \end{align*}

    \item \emph{Baseline Performance}: Opting out of personalization should guarantee the expected performance of a \emph{generic model} $\clf{}: \X \to \Y$ where $\clf \in \argmin_{h \in \Hset}\truerisk{}{h}$. $$\truerisk{\gb}{\psys{\rtree}(\cdot,\Dnr{})} \leq \truerisk{\gb}{\clf} \textrm{ for all groups } \gb \in \G.$$
\end{enumerate}
\label{Def::ParticipatorySystem}
\end{definition}
We denote the gains of personalization for group $\gb$ in terms of true risk and empirical risk as $\truegap{\gb}{\rg}{\Dnr{}}  := \truerisk{\gb}{\psys{\rtree}(\cdot, \Dnr{})} - \truerisk{\gb}{\psys{\rtree}(\cdot, \rg)}$ and $\empgap{\gb}{\rg}{\Dnr{}}  := \emprisk{\gb}{\psys{\rtree}(\cdot, \Dnr{})} - \emprisk{\gb}{\psys{\rtree}(\cdot, \rg)}$. 

Here, the \emph{Incentive Compatibility} property guarantees that opting into personalization will improve performance -- i.e. when individuals provide their personal information, the system can effectively leverage that data to deliver more accurate predictions in expectation.
The \emph{Baseline Performance} property safeguards performance for individuals who opt out of personalization -- i.e., it ensures that individuals who choose not to share their personal information receive the performance of a generic model trained on a dataset without group attributes. 
Together, these properties uphold the principle of \emph{rationality}, in which users always receive gains from personalization such that $\truegap{\gb}{\rg}{\Dnr{}}>0$ for all $\gb \in \G, \rg \in \dim{\gb}$. 

\newcommand{\titlecell}[2]{%
\setlength{\tabcolsep}{0pt}%
\begin{tabular}{p{#1\linewidth}}#2\end{tabular}%
}

\newcommand{\textcell}[2]{%
\setlength{\tabcolsep}{0pt}%
\begin{tabular}{>{\footnotesize\raggedright}p{#1\linewidth}}%
#2%
\end{tabular}%
}
\begin{figure*}[t!]
\centering
\resizebox{\linewidth}{!}{
\begin{tabular}{ccp{0.5\linewidth}}
\cell{c}{Interface} & \cell{c}{Sample Interface} & \cell{c}{Description} \\ 
\cmidrule(lr){1-1}\cmidrule(lr){2-2}\cmidrule(lr){3-3}

\cell{l}{\textsf{Minimal}} &
\cell{c}{\includegraphics[page=1,trim=1.66in 5.2in 9.16in 0.56in,clip,width=0.25\linewidth]{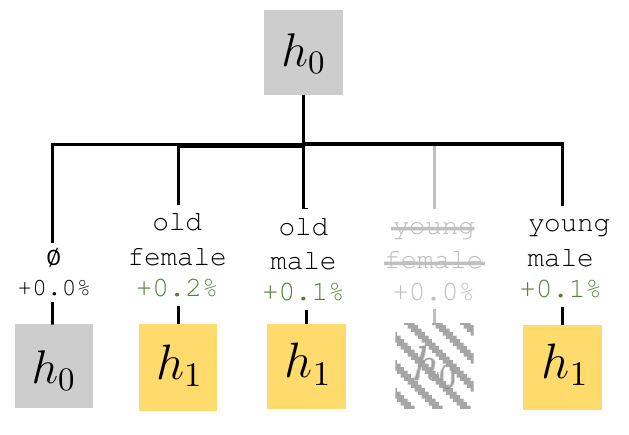}} &
\textcell{1.0}{\emph{Minimal systems} let individuals opt out of personalization from existing personalized model $\plf{}$. Individuals who opt out receive predictions from a generic model $\clf{}$ trained without group attributes. These systems can be built by training one additional model. 
}
\\ \cmidrule(lr){1-1}\cmidrule(lr){2-2}\cmidrule(lr){3-3}

\cell{l}{\textsf{Flat}} & 
\cell{c}{\includegraphics[page=1,trim=5.17in 5.2in 3.95in 0.56in, clip, width=0.45\linewidth]{figures/participatory_systems_figures.pdf}} & 
\textcell{1.0}{%
\emph{Flat systems} support \emph{partial personalization} by allowing each person to report any subset of group attributes. Thus, a person with $\gb_i = [\texttt{old}, \texttt{female}]$ can report $\rg_i =  [\texttt{old}, \dnr]$. These systems can improve performance by using a distinct model to assign personalized predictions to each reporting group.
}
\\  \cmidrule(lr){1-1}\cmidrule(lr){2-2}\cmidrule(lr){3-3}
\cell{c}{\textsf{Sequential}} & 
\cell{c}{\includegraphics[page=1,trim=9.47in 4.34in 0.2in 0.6in, clip, width=0.40\linewidth]{figures/participatory_systems_figures.pdf}}  &
\textcell{1.0}{\emph{Sequential systems} let support partial personalization while allowing individuals to report one attribute at a time. This interface is better suited to inform consent as users can make reporting decisions by comparing two models simultaneously. They are also well-suited for personalization tasks with information that must be acquired at prediction time (e.g., the outcome of a test result).} 
\end{tabular}
}
\caption{Participatory systems for personalized classification task with group attributes $\textgp{sex} \times \textgp{age} = [\textgp{male}, \textgp{female}] \times [\textgp{old}, \textgp{young}]$. Each system allows a person to opt out of personalization by reporting $\dnr$ and informs this choice through the gains of personalization (e.g., \textgp{+0.2\%} gain in accuracy). Systems minimize data use by removing reporting options that do not lead to gain (e.g., $[\textgp{young}, \textgp{female}]$ is pruned in all systems as it leads to a gain $\leq$ 0.0\%. We show reporting options pruned using grey-striped boxes.}
\label{Fig::ParticipatorySystemsGallery}
\end{figure*}

The simplest way to facilitate consent is to impute the group membership of individuals who opt out of personalization at prediction time. Imputation allows individuals to opt out of personalization but does not guarantee the accuracy of predictions for individuals who opt out. Formally, imputation violates the baseline performance guarantee of a participatory system. That is, if users opt out of personalization ($\rg = \Dnr$), they may receive a less accurate prediction than they would have from a generic model $\truerisk{\gb}{\psys{\rtree}(\cdot,\hat{\gb})} > \truerisk{\gb}{\clf}$ where $\hat{\gb}$ denote the imputed attributes. For example, in a setting where we could perfectly impute group membership, a group might be assigned better predictions from a generic model (see \cref{Fig::MinimalParticipation}). In the worst case, imputation may be incorrect, leading to even more inaccurate predictions than those of the generic or personalized model.



\newcommand{\utility}[2]{u_{#1}({#2})}
\newcommand{\cost}[2]{c_{#1}({#2})}
\newcommand{\benefit}[2]{b_{#1}({#2})}

\subsection{Agent Model for Individual Disclosure}
\label{Appendix::ModelAsAMarket}

The performance of participatory systems will depend on individual reporting decisions. In what follows, we characterize how participatory systems will perform under a generalized model of individual disclosure. Given a participatory system $\plf{}: \X \times \RG{} \to \Y$, we assume that each person will report group membership as
$\rg_i \in \argmax_{\rg \in \RG}{\utility{i}{\rg;\plf{}}}$. 
Here, the utility function can be $$\utility{i}{\rg;\plf{}} = \benefit{i}{\rg; \plf{}} - \cost{i}{\rg},$$ where $\cost{i}{\cdot}$ and $\benefit{i}{\cdot}$ denote their cost and benefit of disclosure, respectively. 
We assume that costs increase monotonically with information that is disclosed so that $\cost{i}{\rg} \geq 0$ for all $\rg \in \RG$ and $\cost{i}{\rg} \leq \cost{i}{\rg'}$ for $\rg \subseteq \rg'$. 
We assume that benefits increase monotonically with true risk so that $\benefit{i}{\rg,\plf{}} > \benefit{i}{\rg',\plf{}}$ when $\truerisk{\rg}{\plf{}(\xb_i, \rg)} < \truerisk{\rg}{\plf{}(\xb_i, \rg')}.$

The following remarks apply to any participatory system $\psys{}: \X \times \RG \to Y$ that include a personalized model $\plf{}: \X \times \G \to \Y$ and a generic model $\clf{}: \X \to \Y$ as its components. 

\begin{itemize}[leftmargin=*]

\item Every participatory system $\psys{}$ will perform as well as a generic model $\clf$. When a personalized model $\plf{}$ requires users to report information detrimental to performance (see \cref{Fig::MinimalParticipation}), individuals incur a cost of disclosure without receiving a benefit. In such instances, a minimal system $\psys{}: \X \times \RG^{\min} \to Y$ would allow individuals to opt out of detrimental personalization and receive predictions from a generic model. 

\item Every participatory system $\psys{}$ with more reporting options will perform better. Since utility can only increase with the number of reporting options, the maximum utility for each person will exceed that of a minimal system. Thus, flat and sequential systems will perform better than a minimal system. 

\item The best-case performance of any participatory system will exceed the performance of any of its components. Thus, we are guaranteed that any participatory system will outperform a traditional personalized model so long as it is considered a component. 

\end{itemize}

\subsection{Profiling System Performance with Respect to Participation}

We can use the models for individual disclosure to evaluate how a participatory system will perform once it is deployed. Given a participatory system, we can evaluate this by simulating the parameters in the individual disclosure model shown above. We can then summarize the results from this evaluation for each intersectional group through a performance profile that shows how the system performance will vary across different levels of participation.

%
\begin{figure}[h]
\centering
\textgp{[<30}, \textgp{HIV+]} \hspace{0.3\linewidth} \textgp{[>30}, \textgp{HIV-]}  \\
\includegraphics[width=0.45\linewidth]{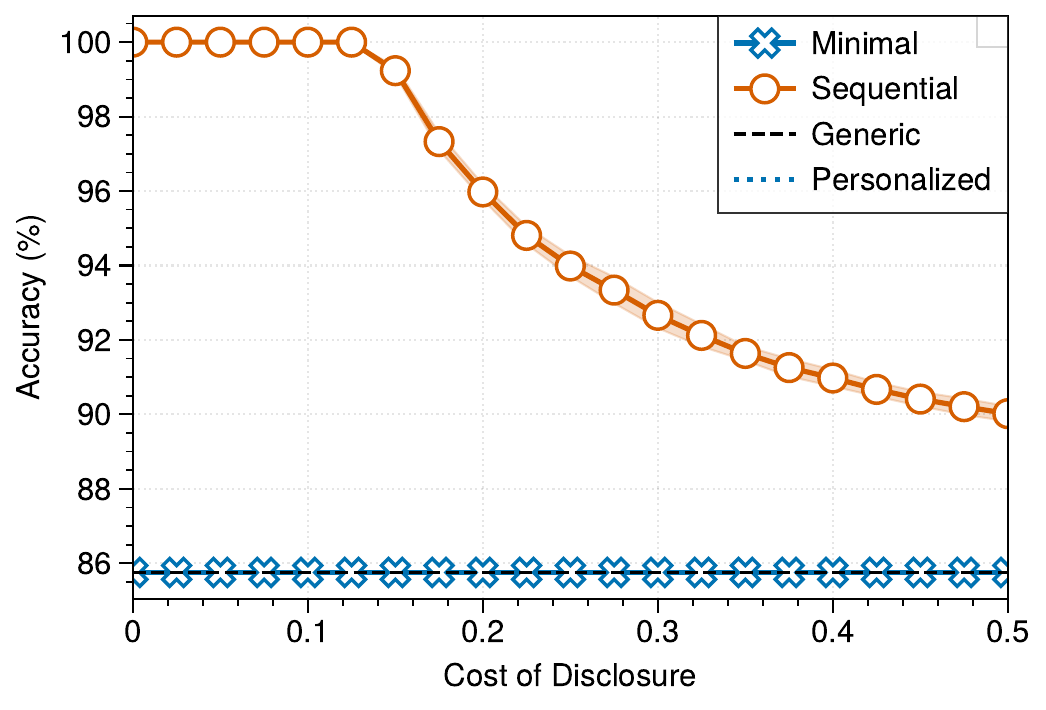}
\includegraphics[width=0.45\linewidth]{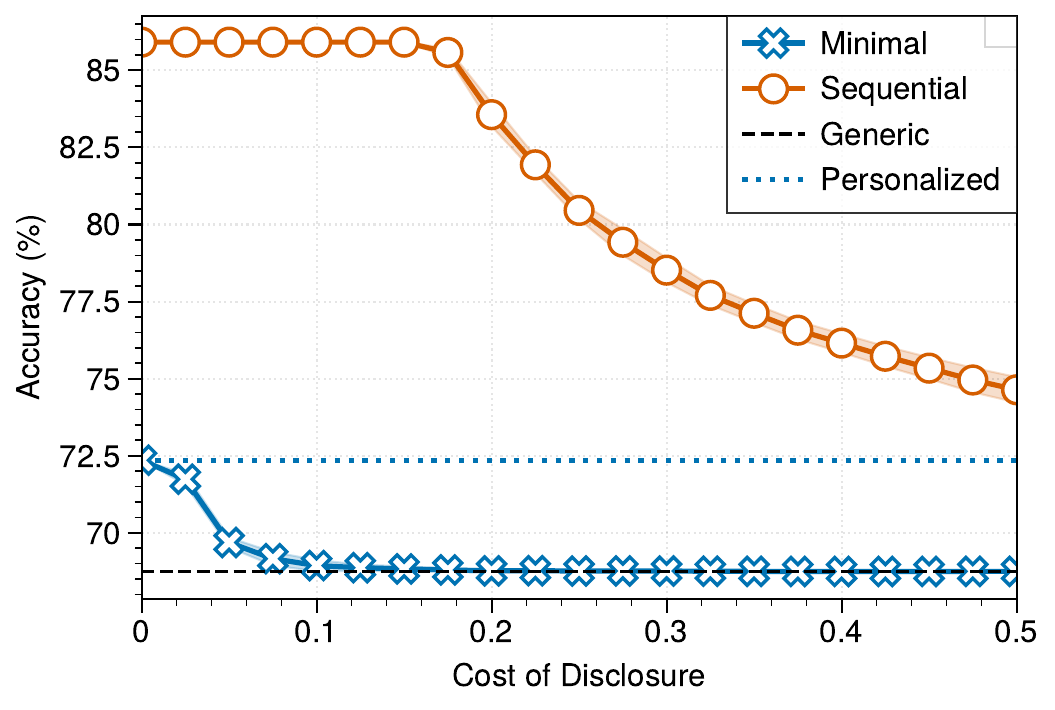} \\
\textgp{[<30}, \textgp{HIV-]} \hspace{0.3\linewidth}  \textgp{[>30}, \textgp{HIV-]}  \\
\includegraphics[width=0.45\linewidth]{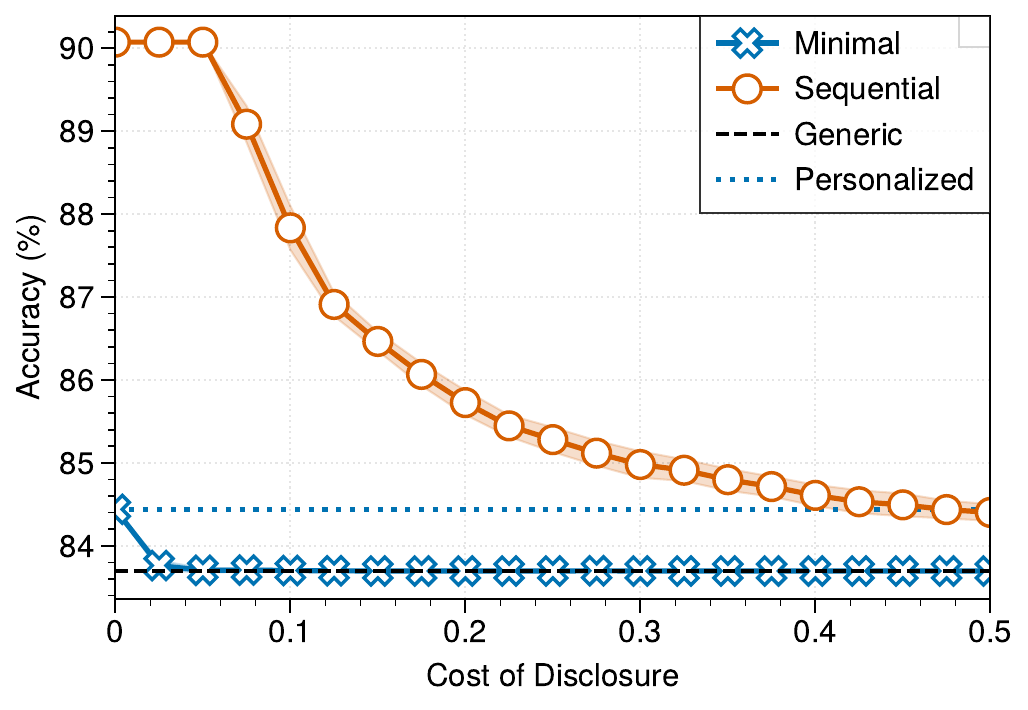}
\includegraphics[width=0.45\linewidth]{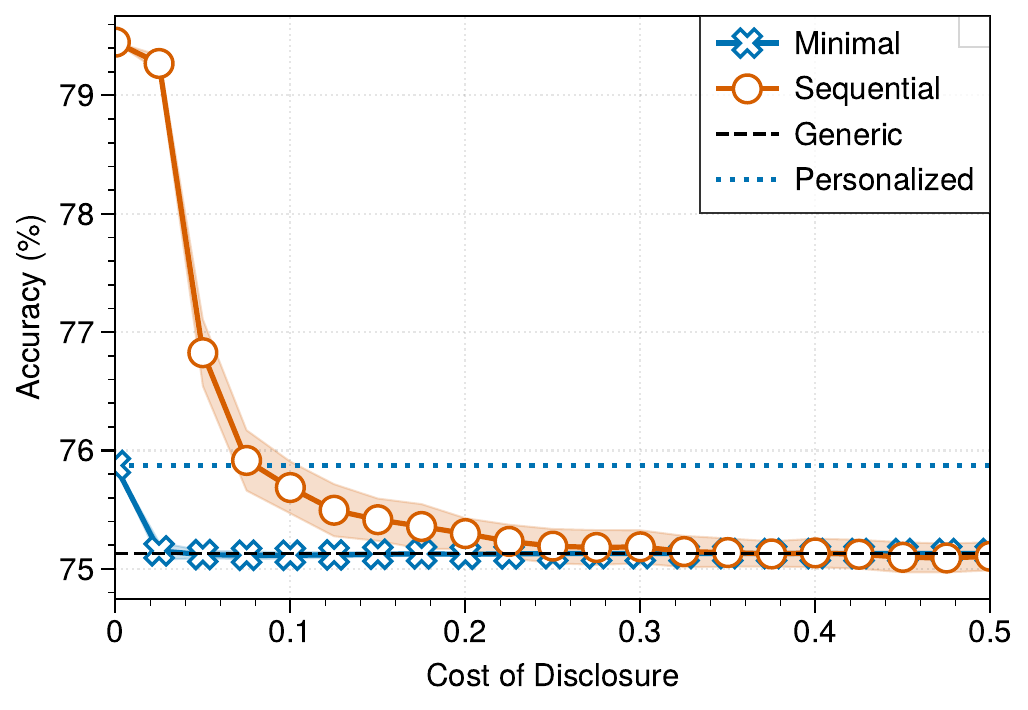}
\caption{Performance profiles of the simulations performed for each intersectional group in the \textds{saps} dataset. The sequential system outperforms static personalized systems when all group attributes are reported. When the cost of reporting is high, the sequential system still outperforms minimally personalized systems, as evidenced by higher accuracy at varying reporting cost thresholds.}
\label{Fig::ParticipationProfile2}
\end{figure}

\section{Learning Participatory Systems}
\label{Sec::Algorithms}
\newcommand{\alltrees}{\mathbb{T}}
\newcommand{\tree}{T}
\newcommand{\node}{\rg{}}
\newcommand{\textprc}[1]{{\small\textsf{#1}}}

This section describes a model-agnostic algorithm to learn participatory systems that ensure incentive compatibility and baseline performance in deployment.
%
%
We outline our procedure in \cref{Alg::ParticipatorySystem} to learn the three kinds of participatory systems in \cref{Fig::ParticipatorySystemsGallery}. The procedure takes as input a pool of candidate models $\candidatepool{}$ and datasets for assignment and pruning $\data{}^\textrm{train}$ and $\data{}^\textrm{valid}$. It outputs a collection of participatory systems that obey the desired properties on test data. The procedure combines three routines to (1) generate viable reporting interfaces (\cref{AlgStep::Enumerate}); (2) assign models over the interface (\cref{AlgStep::Assign}); (3) prune the system to limit unnecessary data collection (\cref{AlgStep::Prune}). We discuss each routine below and present a complete procedure in \cref{Appendix::Algorithms}.

\begin{algorithm}[!h]
\begin{algorithmic}[1]\small
\alginput{$\candidatepool{}: \{h: \X \times \G \to \Y \}$}{pool of candidate models} 
\alginput{$\data{}^\textrm{assign} = \{(\xb_i, \gb_i, y_i)\}_{i=1}^{n^\textrm{assign}}$}{training dataset}
\alginput{$\data{}^\textrm{prune} = \{(\xb_i, \gb_i, y_i)\}_{i=1}^{n^\textrm{prune}}$}{validation dataset}
\State $\alltrees \gets  \textprc{ViableTrees}(\G, \data{}^\textrm{assign})$ \Comment{$|\alltrees| = 1$ for minimal \& flat systems} \label{AlgStep::Enumerate} 
\For{$\tree \in \alltrees{}$}
\State $T \gets \textprc{AssignModels}(T,\candidatepool, \data{}^\textrm{assign})$ \label{AlgStep::Assign} \Comment{assign models}
\State $T \gets \textprc{PruneLeaves}(T, \data{}^\textrm{prune})$ \label{AlgStep::Prune} \Comment{prune models}
\EndFor
\Ensure{$\alltrees{}$, collection of participatory systems}
\end{algorithmic}
\caption{Learning Participatory Systems}
\label{Alg::ParticipatorySystem}
\end{algorithm}

\paragraph{Model Pool}

Our procedure takes as input a \emph{pool of personalized models} $\candidatepool{}$ to assign over a reporting interface.
At a minimum, a pool should contain two models: a personalized model $\plf{}$ for individuals who opt into personalization and a generic model $\clf{}$ for individuals who opt out of personalization.
A single personalized model can perform unreliably across reporting groups due to differences in the data distribution or trade-offs between groups. Using a pool safeguards against these effects by drawing on models from different model classes that have been personalized using different techniques for each reporting group. By default, we include a model for each reporting group trained using only data for that group, as such models can perform well on heterogeneous subgroups~\cite{ustun2019fairness,suriyakumar2022fairuse}. 

\paragraph{Enumerating Reporting Interfaces}

We call the $\textprc{ViableTrees}$ routine in \cref{AlgStep::Enumerate} to enumerate \emph{viable} reporting interfaces. We only call this routine for sequential systems since $\alltrees{}$ consists of a single tree known a priori for minimal and flat systems.
$\textprc{ViableTrees}$ takes as input a group attributes $\G$ and a dataset $\data{}^\textrm{assign}$. It returns all $m$-ary trees that obey constraints on sample size and reporting (e.g., users who report \textgp{male} should report \textgp{age} before \textgp{HIV}). By default, we only generate trees so that we have sufficient data to estimate gains at each node of the reporting interface (e.g., trees whose leaves contain at least one positive sample, one negative sample, and $\n{\rg} \geq d+1$ samples to avoid overfitting). 
In general, $\textprc{ViableTrees}$ scales to tasks with $\leq 8$ group attributes. Beyond this limit, one can reduce the enumeration size by specifying ordering constraints or a threshold number of trees to enumerate before stopping. For a task with three binary group attributes, $\alltrees{}$ contains $24$ 3-ary trees of depth 3. Given a complete ordering of all $3$ group attributes, $\alltrees$ would have $1$ tree. We can also consider a greedy algorithm (see Appendix \ref{Appendix::GreedyAlgorithm}), which may be practical for large-scale problems.

\paragraph{Model Assignment}
We assign each reporting group a model using the $\textprc{AssignModels}$ routine in \cref{AlgStep::Assign}. 
Given a reporting group $\rg$, we consider all models that could use group membership $\rg_i \in \dim{\rg}$. Thus, a group that reports \textgp{age} and \textgp{sex} could be assigned predictions from a model that requires \textgp{age}, \textgp{sex}, both, or neither. This implies that we can always assign the generic model to any reporting group, ensuring that the model at each node performs as well as the generic model on out-of-sample data (i.e., \emph{baseline performance} in Definition \ref{Def::ParticipatorySystem}).
%

\paragraph{Pruning}



$\textprc{ViableTrees}$ may output trees that violate incentive compatibility by requesting personal information that does not reliably improve performance.
%
This can happen when a system assigns the same model or a model that performs equally well to nested reporting groups -- see, e.g., \cref{Fig::ParticipatorySystemsGallery} where the Flat system assigns $\clf{}$ to $[\textgp{female},\dnr{}]$ and $[\textgp{female}, \textgp{young}]$.
%
The $\textprc{Prune}$ routine in \cref{AlgStep::Prune}. This routine takes as input a participatory system $\psys{\rtree}$ and a validation sample $\data{}^\textrm{prune}$ and outputs a system $\psys{\tree'}$ with a pruned interface $\rtree' \subseteq \tree$ that will not request data in such cases. The routine uses a bottom-up pruning procedure that calls a one-sided hypothesis test at each node:
\begin{align*}
H_0: \truegap{\rg}{\rg}{\parent{\rg}} \leq 0 \qquad H_A: \truegap{\rg}{\rg}{\parent{\rg}} > 0
\end{align*}
The test checks if each reporting group $\rg$ receives more accurate predictions using the personalized model assigned by the current interface $\rtree$ in comparison to its parent $\psys{\rtree}({\cdot,\parent{\rg}})$. Here, $H_0$ assumes a reporting group prefers the parent model. Thus, we reject $H_0$ when evidence suggests that $\psys{\rtree}({\cdot,\rg})$ performs better for $\rg$ on pruning data. The exact test should be chosen based on the performance metric for the underlying prediction task. In general, we can use a bootstrap hypothesis test~\citep[][]{diciccio1996bootstrap} and draw on more powerful tests for salient performance metrics~\citep[e.g.,][for accuracy and AUC]{dietterich1998approximate,delong1988comparing, fastdelong}.

\paragraph{On Computation}

Our approach provides several options for learning participatory systems. For example, one can train only two models and build a minimal system or train a flat or sequential system with a limited number of models in the pool. Nevertheless, the primary bottleneck when learning participatory systems is \emph{data} rather than \emph{compute}. Given a finite sample dataset, we are limited in the number of categorical attributes used for personalization. This is because we require a minimum number of samples for each intersectional group to train a personalized model and evaluate its performance. Given that the number of intersectional groups increases exponentially with each attribute, we quickly enter a regime where we cannot train models for a given group (e.g., because we lack sufficient labels) or reliably evaluate its gain for assignment and pruning~\citep[see][]{paes2022on}.



\section{Experiments}
\label{Sec::Experiments}

\newcommand{\OneHot}[0]{\textmn{1Hot}}
\newcommand{\IntHot}[0]{\textmn{mHot}}
\newcommand{\OneHotImpute}[0]{\textmn{SimpleImpute}}
\newcommand{\OneHotKNNImpute}[0]{\textmn{KNN-1Hot}}
\newcommand{\IntHotKNNImpute}[0]{\textmn{KNN-mHot}}
\newcommand{\DCP}[0]{\textmn{DCP}}
\newcommand{\MinimalSys}[0]{\textmn{Minimal}}
\newcommand{\FlatSys}[0]{\textmn{Flat}}
\newcommand{\SeqSys}[0]{\textmn{Seq}}

We benchmark participatory systems on real-world clinical prediction tasks. Our goal is to evaluate these approaches in terms of performance, data usage, and consent in applications where individuals have a low reporting cost. We include code to reproduce these results in an ~\href{https://github.com/hljames/participatory_systems}{public repository}.

\subsection{Setup}
\label{Sec::ExperimentalSetup}

\newcommand{\adddatainfo}[6]{\cell{l}{%
\textds{#1}\\%
$n={#2}, d={#3}$\\
$\G=\{#5\}$\\
$|\G|={#4}$ groups\\
{#6}
}}

\newcommand{\adddatainfogp}[6]{\cell{l}{%
\textds{#1}\\%
$n={#2}, d={#3}$\\
{#5}\\
{#6}
}}

\newcommand{\apnea}[0]{\adddatainfo{apnea}{1152}{26}{6}{\textgp{age},\textgp{sex}}{\citet[][]{ustun2016clinical}}}
\newcommand{\cshockRmimic}[0]{\adddatainfo{cardio\_mimic}{5289}{49}{8}{\textgp{age},\textgp{sex}, \textgp{race}}{\citet{johnson2016mimic}}}
\newcommand{\cshockReicu}[0]{\adddatainfo{cardio\_eicu}{1341}{49}{8}{\textgp{age}, \textgp{sex}, \textgp{race}}{\citet{pollard2018eicu}}}
\newcommand{\seercrcagesexcalifornia}[0]{\adddatainfo{coloncancer}{29211}{72}{6}{\textgp{age}, \textgp{sex}}{\citet[][]{nci2019seer}}}
\newcommand{\seerrespacmcollapsedagesexall}[0]{\adddatainfo{lungcancer}{120641}{84}{6}{\textgp{age}, \textgp{sex}}{\citet[][]{nci2019seer}}}
\newcommand{\support}[0]{\adddatainfo{support}{9105}{55}{6}{\textgp{age}, \textgp{sex}}{\citet[][]{knaus1995support}}}
\newcommand{\saps}[0]{\adddatainfo{saps}{7797}{36}{4}{\textgp{HIV},\textgp{age}}{\citet[][]{allyn2016simplified}}}



We consider six classification tasks for clinical decision support where we personalize a model with group attributes that are protected or sensitive (see \cref{Table::Results} and \cref{Appendix::Datasets}). %
Each task pertains to an application where we would expect individuals to have a low cost of reporting -- and therefore report personal information so long as there is any benefit. This is because the information used for personalization is readily available, relevant to the prediction task, and likely to be disclosed given laws surrounding the privacy and confidentiality of health data~\citep{hipaa}. \citep{bansal2010impact,anderson2011digitization}.
One exception is \textds{cardio\_eicu} and $\textds{cardio\_mimic}$, which are personalized based on race and ethnicity.
%
We note that the use of race in clinical risk scores should be approached with caution \cite{vyas2020hidden}; participatory systems offer one way to safeguard against inappropriate use.
%
%
%
We split each dataset into a test sample (20\% for evaluating out-of-sample performance) and a training sample (80\% for training, pruning, assignment, and estimating gains to show users). We train three kinds of personalized models for each dataset:
\begin{itemize}
\item \emph{Static}: These models are personalized using a one-hot encoding of group attributes (\OneHot{}) and a one-hot encoding of intersectional groups (\IntHot) 
\item \emph{Imputed}: These are variants of static models where we impute the values of group attributes (\OneHotKNNImpute{}, \IntHotKNNImpute). In practice, the performance of imputation will range between the values reported for \OneHot{} and $\IntHotKNNImpute{}$ (100\% opt-in) and \OneHotKNNImpute{}, \IntHotKNNImpute (100\% opt-out).

\item \emph{Participatory}: These are participatory systems built using our approach. These include \MinimalSys{}, a minimal system built from \OneHot{} and its generic counterpart; and \FlatSys{} and \SeqSys{}, flat and sequential systems built from \OneHot{}, \IntHot~ and their generic counterparts.
\end{itemize}
We train all models -- personalized models and the components of participatory systems -- from the same model class and evaluate them using the metrics in \cref{Table::Metrics}. We repeat the experiments four times, varying the model class (logistic regression, random forests) and the salient performance metric (error rate for decision-making and AUC for ranking) to evaluate the sensitivity of our findings with respect to model classes and classification tasks.

\newcommand{\requested}[1]{\textrm{requested}({#1})}
\begin{table*}[h!]
\centering
\resizebox{\linewidth}{!}{
\footnotesize
    \begin{tabular}{@{\,}>{\raggedright\arraybackslash}p{0.16\linewidth}@{\;}>{\raggedright\arraybackslash\scriptsize}p{0.16\linewidth}>{\footnotesize}p{0.625\linewidth}@{\,}}
\textheader{Metric} & 
{\footnotesize\textheader{Definition}} & 
{\footnotesize\textheader{Description}}%
\\ \midrule
Overall Performance & 
$\displaystyle\sum_{\gb\in\G} \tfrac{\n{\gb}}{n}\emprisk{\gb}{\plf{\gb}}$ & Population-level performance of a personalized system/model, computed as a weighted average over all groups
\\ \midrule
Overall Gain & 
$\displaystyle\sum_{\gb \in \G} \tfrac{\n{\gb}}{n} \empgap{\gb}{\gb}{\Dnr{}}$ &
{Population-level gain in performance of a personalized system/model over its generic counterpart}
\\ \midrule
Group Gains & 
$\displaystyle \min_{\gb \in \G}/\max_{\gb \in \G} \empgap{\gb}{\gb}{\Dnr{}}$ &
{Range of gains of a personalized system/model over its generic counterpart across all groups}
\\ \midrule
Rationality Violations & 
$\displaystyle \sum_{\gb \in \G} \mathbb{I}[\textrm{reject}~H_0]$ &   
Number of rationality violations detected using a bootstrap hypothesis test with 100 resamples and significance 
level of 10\% where $H_0:\truegap{\gb}{\gb}{\Dnr{}} \geq 0$ and $H_A:\truegap{\gb}{\gb}{\Dnr{}}<0$%
\\ \midrule
Imputation Risk & 
$\displaystyle \min_{\gb \in \G}\empgap{\gb}{\gb}{\gb'}$ &   
Worst-case loss in performance due to a result of incorrect imputation. This measure can only be computed for static models only.
\\ \midrule
Options Pruned & 
$\displaystyle \frac{|\RG| - |\RG(\plf{})|}{|\RG|}$ &   
The number of reporting options pruned from a system/model. Here, $\RG(\plf{})$ denotes the options available after $\plf{}$ has been pruned, and $\RG$ denotes options available before pruning. 
\\ \midrule
Data Use &
$\displaystyle \sum_{\gb \in \G} \tfrac{\n{\gb}}{n} \tfrac{\requested{\plf{},\gb}}{\textrm{dim}(\G)}$ &   
Proportion of group attributes $k$ requested by $\plf{}$ from each group, averaged over all groups in $\G$ 

\end{tabular}%
}
\caption{Metrics used to evaluate performance, data use, and consent. We report performance on a held-out test sample. We assume that individuals report group membership to static models, do not report group membership to imputed models, and only report to participatory systems when reporting leads to a positive gain. In the latter case, we estimate the gain shown to individuals using a validation set in the training sample.}
\label{Table::Metrics}
\end{table*}

\subsection{Results}
\label{Sec::ExperimentalResults}

We show results for logistic regression models and error rate in \cref{Table::Results} and for other model classes and classification tasks in \cref{Appendix::ExtraResults}. In what follows, we discuss these results. 

\begin{table*}[!t]
\centering
\resizebox{\linewidth}{!}{
\centering\begingroup\fontsize{8}{10}\selectfont
\renewcommand{\metricsguide}[0]{\cell{r}{Overall Performance\\Overall Gain\\Group Gains\\Rat. Violations\\Imputation Risk\\Options Pruned\\Data Use}}
\begin{tabular}{lrlllllll}
\multicolumn{2}{c}{ } & \multicolumn{2}{c}{\textsc{Static}} & \multicolumn{2}{c}{\textsc{Imputed}} & \multicolumn{3}{c}{\textsc{Participatory}} \\
\cmidrule(l{3pt}r{3pt}){3-4} \cmidrule(l{3pt}r{3pt}){5-6} \cmidrule(l{3pt}r{3pt}){7-9}
Dataset & Metrics & \OneHot{} & \IntHot{} & \OneHotKNNImpute{} & \IntHotKNNImpute{} & \MinimalSys{} & \FlatSys{} & \SeqSys{}\\
\midrule
\apnea & \metricsguide{} & \cell{r}{29.1\%\\0.1\%\\-1.1\% -- 1.2\%\\\badvalue{1}\\-4.9\%\\0/6\\100.0\%} & \cell{r}{29.3\%\\-0.1\%\\-0.8\% -- 0.4\%\\\badvalue{1}\\-5.2\%\\0/6\\100.0\%} & \cell{r}{29.0\%\\0.2\%\\-1.1\% -- 1.2\%\\\badvalue{1}\\\\0/12\\0.0\%} & \cell{r}{27.9\%\\1.3\%\\-0.8\% -- 0.4\%\\\badvalue{1}\\\\0/12\\0.0\%} & \cell{r}{28.9\%\\0.3\%\\0.0\% -- 1.2\%\\0\\\\4/7\\33.3\%} & \cell{r}{\cellcolor{fgood}\textbf{24.1\%}\\\cellcolor{fgood}\textbf{5.1\%}\\0.0\% -- 13.8\%\\0\\\\5/12\\83.3\%} & \cell{r}{24.3\%\\4.9\%\\-0.4\% -- 13.8\%\\0\\\\6/12\\58.3\%}\\
\midrule

\cshockReicu & \metricsguide{} & \cell{r}{21.4\%\\0.4\%\\-1.3\% -- 2.6\%\\\badvalue{1}\\-4.6\%\\0/8\\100.0\%} & \cell{r}{21.5\%\\0.3\%\\-2.7\% -- 3.0\%\\\badvalue{1}\\-5.4\%\\0/8\\100.0\%} & \cell{r}{21.6\%\\0.3\%\\-1.3\% -- 2.6\%\\\badvalue{1}\\\\0/27\\0.0\%} & \cell{r}{22.1\%\\-0.2\%\\-2.7\% -- 3.0\%\\\badvalue{1}\\\\0/27\\0.0\%} & \cell{r}{21.6\%\\0.3\%\\0.0\% -- 2.6\%\\0\\\\6/9\\25.0\%} & \cell{r}{\cellcolor{fgood}\textbf{10.2\%}\\\cellcolor{fgood}\textbf{11.7\%}\\3.1\% -- 20.9\%\\0\\\\10/27\\100.0\%} & \cell{r}{\cellcolor{fgood}\textbf{10.2\%}\\\cellcolor{fgood}\textbf{11.7\%}\\3.1\% -- 20.9\%\\0\\\\9/27\\83.3\%}\\
\midrule

\cshockRmimic & \metricsguide{} & \cell{r}{19.4\%\\-0.1\%\\-0.9\% -- 0.4\%\\\badvalue{3}\\-1.1\%\\0/8\\100.0\%} & \cell{r}{19.3\%\\-0.0\%\\-0.9\% -- 0.5\%\\\badvalue{2}\\-1.1\%\\0/8\\100.0\%} & \cell{r}{19.3\%\\-0.0\%\\-0.9\% -- 0.4\%\\\badvalue{3}\\\\0/27\\0.0\%} & \cell{r}{20.1\%\\-0.8\%\\-0.9\% -- 0.5\%\\\badvalue{2}\\\\0/27\\0.0\%} & \cell{r}{19.2\%\\0.1\%\\0.0\% -- 0.4\%\\0\\\\6/9\\25.0\%} & \cell{r}{\cellcolor{fgood}\textbf{15.7\%}\\\cellcolor{fgood}\textbf{3.5\%}\\-1.6\% -- 9.8\%\\\badvalue{1}\\\\6/27\\100.0\%} & \cell{r}{\cellcolor{fgood}\textbf{15.7\%}\\\cellcolor{fgood}\textbf{3.5\%}\\-1.6\% -- 9.8\%\\\badvalue{1}\\\\8/27\\91.7\%}\\
\midrule

\seercrcagesexcalifornia & \metricsguide{} & \cell{r}{37.0\%\\0.1\%\\-0.4\% -- 0.3\%\\\badvalue{1}\\-1.4\%\\0/6\\100.0\%} & \cell{r}{36.7\%\\0.4\%\\-0.1\% -- 1.1\%\\0\\-0.9\%\\0/6\\100.0\%} & \cell{r}{37.0\%\\0.1\%\\-0.4\% -- 0.3\%\\\badvalue{1}\\\\0/12\\0.0\%} & \cell{r}{36.9\%\\0.2\%\\-0.1\% -- 1.1\%\\0\\\\0/12\\0.0\%} & \cell{r}{37.0\%\\0.1\%\\0.0\% -- 0.3\%\\0\\\\5/7\\16.7\%} & \cell{r}{36.6\%\\0.5\%\\0.0\% -- 1.7\%\\0\\\\7/12\\50.0\%} & \cell{r}{\cellcolor{fgood}\textbf{36.1\%}\\\cellcolor{fgood}\textbf{1.0\%}\\0.2\% -- 1.7\%\\0\\\\5/12\\75.0\%}\\
\midrule

\seerrespacmcollapsedagesexall & \metricsguide{} & \cell{r}{19.6\%\\-0.1\%\\-0.4\% -- 0.2\%\\\badvalue{4}\\-0.5\%\\0/6\\100.0\%} & \cell{r}{19.6\%\\-0.1\%\\-0.3\% -- 0.2\%\\\badvalue{4}\\-0.5\%\\0/6\\100.0\%} & \cell{r}{19.9\%\\-0.3\%\\-0.4\% -- 0.2\%\\\badvalue{4}\\\\0/12\\0.0\%} & \cell{r}{19.8\%\\-0.2\%\\-0.3\% -- 0.2\%\\\badvalue{4}\\\\0/12\\0.0\%} & \cell{r}{19.5\%\\0.0\%\\0.0\% -- 0.0\%\\0\\\\6/7\\0.0\%} & \cell{r}{18.9\%\\0.6\%\\0.0\% -- 0.9\%\\0\\\\3/12\\83.3\%} & \cell{r}{\cellcolor{fgood}\textbf{18.9\%}\\\cellcolor{fgood}\textbf{0.6\%}\\0.3\% -- 0.9\%\\0\\\\7/12\\58.3\%}\\
\midrule

\saps & \metricsguide{} & \cell{r}{20.4\%\\1.3\%\\0.0\% -- 3.6\%\\0\\0.0\%\\0/4\\100.0\%} & \cell{r}{20.7\%\\1.0\%\\0.0\% -- 2.7\%\\0\\-2.4\%\\0/4\\100.0\%} & \cell{r}{20.4\%\\1.3\%\\0.0\% -- 3.6\%\\0\\\\0/9\\0.0\%} & \cell{r}{29.4\%\\-7.7\%\\0.0\% -- 2.7\%\\0\\\\0/9\\0.0\%} & \cell{r}{20.4\%\\1.3\%\\0.0\% -- 3.6\%\\0\\\\1/5\\75.0\%} & \cell{r}{\cellcolor{fgood}\textbf{11.1\%}\\\cellcolor{fgood}\textbf{10.6\%}\\4.3\% -- 17.2\%\\0\\\\1/9\\100.0\%} & \cell{r}{11.1\%\\10.6\%\\4.3\% -- 17.2\%\\0\\\\3/9\\75.0\%}

\end{tabular}
\endgroup{}

}
\caption{Participatory systems and personalized models for all datasets. We summarize metrics in \cref{Table::Metrics} and present results for other model classes and prediction tasks in \cref{Appendix::ExtraResults}.}
\label{Table::Results}
\end{table*}


\paragraph{On Performance} 

Our results in \cref{Table::Results} show that participatory systems can improve performance across reporting groups. Here, \FlatSys{} and \SeqSys{} achieve the best overall performance on 6/6 datasets and improve the gains from personalization for every reporting group on 5/6 datasets. In contrast, traditional models improve overall performance while reducing performance at a group level (see rationality violations on five datasets for \OneHot{}, \IntHot{}). 
The performance benefits from participatory systems stem from (i) allowing users to opt out of these instances of ``worsenalization'' and (ii) assigning personalized predictions with multiple models. Using \cref{Table::Results}, we can measure the impact of (i) by comparing the performance of \MinimalSys{} vs. \OneHot{}, and the impact of (ii) by comparing the performance of \MinimalSys{} to \FlatSys{} (or \SeqSys{}). For example, on \textds{apnea}, \OneHot{} exhibits a significant rationality violation for group $[\textgp{30\_to\_60}, \textgp{male}]$, meaning they would have been better off with a generic model that did not use their personal data. By comparing the performance of \OneHot{} to \MinimalSys{}, we see that allowing users to opt out of worsenalization reduces test error from 29.1\% to 28.9\%. By comparing the performance on \MinimalSys{} to \FlatSys{} and \SeqSys{}, we see that using multiple models can further reduce test error from 28.9\% to 24.1\%.




\paragraph{On Informed Consent}

\begin{figure*}[!t]
\centering
\resizebox{\linewidth}{!}{
\includegraphics[page=3,trim=.89in 3.03in 0.1in 1.56in, clip,width=\linewidth]{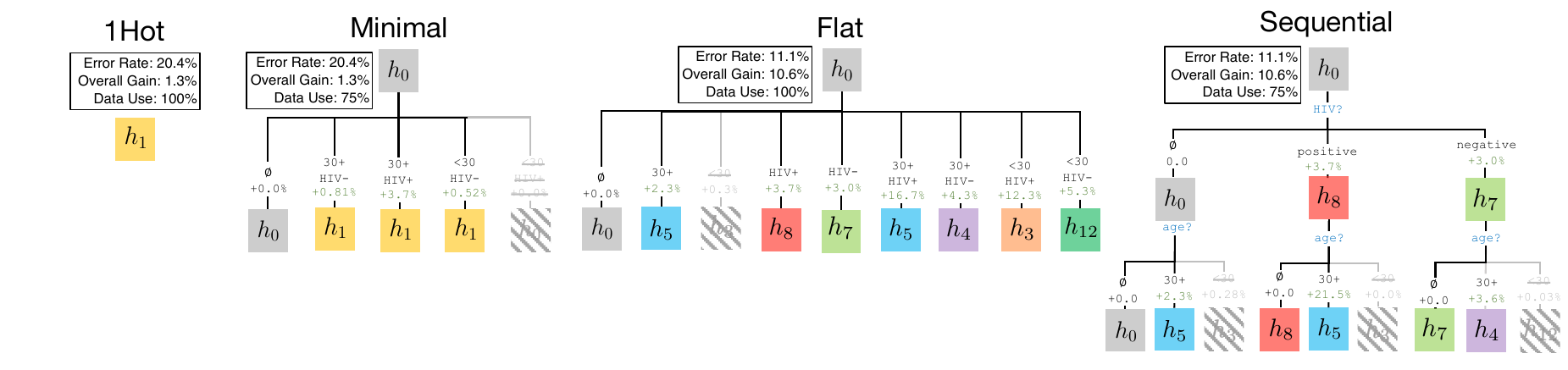}
}
\caption{Participatory systems for the \textds{saps} dataset. These models predict ICU mortality for groups defined by $\G = \textgp{HIV} \times \textgp{age} = \textgp{[+,-]} \times \textgp{[<30, 30+]}$. Here, $\clf{}$ is a generic model, $\plf{1}$ is a \OneHot{} model fit with a one-hot encoding of $\G$, and $h_2 \cdots h_m$ are \OneHot{} and \IntHot{} models fit for each reporting group. We show the gains of each reporting option above each box, and highlight pruned options in grey. For example, in \SeqSys{}, group \textgp{(HIV+, 30+)} sees an estimated $21.5\%$ error reduction for \textgp{age} after reporting \textgp{HIV}. In contrast, group \textgp{(HIV+, <30)} sees no gain from reporting \textgp{age} in addition to \textgp{HIV} status, so this option is pruned.}
\label{Fig::ParticipatorySystemsSaps}
\end{figure*}

Our results show how \FlatSys{} and \SeqSys{} systems can inform consent by allowing users to report a subset of group attributes (e.g., by including reporting options such as $[\textgp{30+}, \dnr{}]$ or $[\dnr{}, \textgp{HIV+}]$). 
Although both \FlatSys{} and \SeqSys{} systems allow for partial personalization, their capacity to inform consent differs.
In a flat system, users may inaccurately gauge the marginal benefit of reporting an attribute by comparing the gains between reporting options. For example, in \cref{Fig::ParticipatorySystemsSaps}, users who are HIV positive would see a gain of $3.7\%$ for reporting $[\dnr{}, \textgp{HIV+}]$, and $16.7\%$ for reporting $[\textgp{30+}, \textgp{HIV+}]$ and may mistakenly conclude that the gain of reporting $\textgp{age}$ is $16.7\% - 3.7\% = 13.0\%$. This estimate incorrectly presumes that the gains of $3.7\%$ were distributed equally across age groups.
Sequential systems directly inform users of the gains for partial reporting. In the sequential system, group $[\textgp{30+}, \textgp{HIV+}]$ is informed that they would see a marginal gain of 21.5\% for reporting $\textgp{age}$, while group $[\textgp{<30}, \textgp{HIV+}]$ is informed they would see a marginal gain of reporting $\textgp{age}$ of $0.0\%$.

\paragraph{On Data Minimization}

Our results show that participatory systems perform better across all groups while requesting less personal data on 6/6 datasets. For example, on \textds{cardio\_eicu}, \SeqSys{} reduces error by 
$11.3\%$ compared to \OneHot{} while requesting, on average, 
$83.3\%$ of the data needed by \OneHot{}. 
In general, participatory systems can limit data use where personalization does not improve performance, e.g., on \textds{lungcancer}.
Even as attributes like \textgp{sex} or \textgp{age} may be readily reported by patients for any performance benefit, limiting data use is valuable when there is a tangible cost associated with data collection -- e.g., when models make use of rating scale for a mental disorder that must be administered by a clinician \citep[][]{sharp2015hamilton}.
The potential for data minimization varies substantially across prediction tasks. On \textds{apnea}, for example, we can prune 6 reporting options when building a \SeqSys{} for decision making (which optimizes error) but 4 options for \SeqSys{} for ranking (which optimizes AUC; see~\cref{Appendix::LogRegAuc}). Overall, participatory systems satisfy ``global data minimization'' as proposed in \cite{biega2020operationalizing}, in that they minimize the amount of per-user data requested
while achieving the quality of a system with access to the full data
on average.


\paragraph{On the Benefits of a Model-Agnostic Approach}

Our findings highlight some of the benefits of a model-agnostic approach, in which we can draw on a rich set of models to achieve better performance while mitigating harm. The resulting system can balance training costs with performance benefits. For example, on the \textds{cardio\_eicu} dataset, training a pool of $126$ candidate models takes 6.5 seconds for logistic regression and 27.5 seconds for random forests on a single CPU.
We can also ensure generalization across reporting groups -- e.g., by a generic model fit from a complex model class, and personalized models fit from a simpler model class. As expected, fitting for a complex model class can lead to considerable changes in overall accuracy -- e.g., we can reduce overall test error for a personalized model from 20.4\% to 14.1\% on \textds{saps} by fitting a random forest rather than a logistic regression model (see \cref{Appendix::ExtraResults}). However, a gain in overall performance does not always translate to gains at the group level. On \textds{saps}, for example, using a random forest also introduces a rationality violation for one group.

\paragraph{On the Pitfalls of Imputation}

One of the simplest approaches to allow individuals to opt out of personalization is to pair a personalized model with an imputation technique. Although this approach can facilitate consent, it may violate the requirements in \ref{Sec::ProblemStatement}. Consider a personalized model that exhibits ``worsenalization'' in \cref{Fig::MinimalParticipation}. Even if one could correctly impute the group membership for every person, individuals may receive more accurate predictions from a generic model $\clf{}$. In practice, imputation is imperfect -- as individuals who opt out of reporting their group membership to a personalized model may be assigned ``worse'' predictions because they are imputed the group membership of a different group. In such cases, opting out may be beneficial, making it difficult for model developers to promote participation while informing consent.
Our results highlight the prevalence of these effects across in practice. For example, on \textds{cardio\_eicu} the estimated ``risk of imputation'' is 
$-4.6\%$
, indicating that every intersectional group can experience an additional 
$4.6\%$ 
error rate as a result of incorrect imputation. The results for \OneHotKNNImpute{} show that this estimate loss in performance can be realized in practice using KNN-imputation, as we find that the imputed system leads to rationality violations on 5/6 datasets. 

\section{Limitations and Future Work}
\label{Sec::ConcludingRemarks}

In this work, we introduced a family of prediction models that allow individuals to report personal data at prediction time. These systems can inform consent while producing large improvements in performance and data use for each group that reports personal data. 

\paragraph{Limitations}
One limitation of this work includes the ability to retrain models on data collection during deployment. This setting arises because data collected post-deployment could exhibit ``non-response'' bias as individuals opt out, and it would be misleading to use it to refine estimates or retrain models. If we retrain using only data from deployment, it is possible and even likely to lose granularity for evaluating certain groups. This setting, however, may violate the principle of \emph{purpose specification} in data collection~\cite{oecd2013}. If the purpose of data collection is to monitor or improve a model, then individuals could be asked to report information voluntarily for this purpose. If data collection aims to improve the model, however, then individuals are within their right to opt-out and should, at minimum, not be harmed by opting in. Solutions could include asking individuals to report their data to improve the system voluntarily or reweighting data collected under non-response bias by comparing the prevalence of groups in deployment to the prevalence in the baseline dataset (though this is an open problem).
In addition, reaping the benefits of personalization hinges on the ability to effectively inform users of the impacts of disclosure~\cite{anderson2011digitization}. Participatory systems are designed to not impose restrictions on what we would present to users – specifically because the information we’d want to show to inform consent should change across applications and target audiences to inform consent~\citep[][]{edwards2013personalised}. There are good ``default practices'' that avoid pitfalls that would most affect a large collection of end-users (i.e., the kind of topic that would be pursued in a user study). In general, we can draw on a broad set of literature in numeracy and uncertainty quantification to promote understanding among end-users~\citep[][]{estrada1999health,nanayakkara2022visualizing}.

\paragraph{Future Work}
One exciting avenue of future research is to evaluate participatory systems in the context of user studies. While this work lays the foundation for building and evaluating participatory systems, future research should include work on the user interface design of participatory systems (e.g., how to communicate gains) and evaluation of these choices (e.g., how well do users choices align with their personal costs and benefits of data provision and performance). 
Another avenue of future work could including exploring participatory systems in the context of continuous attributes and unstructured data. The key requirement for applying this framework to other settings is the ability to estimate the gains of personalization. We could estimate the gains in this work because we had ``groups.'' When tasked with unstructured data, we may be able to accomplish this when we have multiple observations per person (e.g., in speech recognition). On the other hand, continuous attributes may prove more difficult and may require a different approach.


\clearpage
\small
\bibliographystyle{ext/icml2023}
\bibliography{participatory_classification}

\begin{thebibliography}{64}
\providecommand{\natexlab}[1]{#1}
\providecommand{\url}[1]{\texttt{#1}}
\expandafter\ifx\csname urlstyle\endcsname\relax
  \providecommand{\doi}[1]{doi: #1}\else
  \providecommand{\doi}{doi: \begingroup \urlstyle{rm}\Url}\fi

\bibitem[Abajian et~al.(2018)Abajian, Murali, Savic, Laage-Gaupp, Nezami,
  Duncan, Schlachter, Lin, Geschwind, and Chapiro]{abajian2018predicting}
Abajian, A., Murali, N., Savic, L.~J., Laage-Gaupp, F.~M., Nezami, N., Duncan,
  J.~S., Schlachter, T., Lin, M., Geschwind, J.-F., and Chapiro, J.
\newblock Predicting treatment response to intra-arterial therapies for
  hepatocellular carcinoma with the use of supervised machine learning---an
  artificial intelligence concept.
\newblock \emph{Journal of Vascular and Interventional Radiology}, 29\penalty0
  (6):\penalty0 850--857, 2018.

\bibitem[Agresti(2018)]{agresti2018introduction}
Agresti, A.
\newblock \emph{An introduction to categorical data analysis}.
\newblock John Wiley \& Sons, 2018.

\bibitem[Allyn et~al.(2016)Allyn, Ferdynus, Bohrer, Dalban, Valance, and
  Allou]{allyn2016simplified}
Allyn, J., Ferdynus, C., Bohrer, M., Dalban, C., Valance, D., and Allou, N.
\newblock Simplified acute physiology score ii as predictor of mortality in
  intensive care units: a decision curve analysis.
\newblock \emph{PloS one}, 11\penalty0 (10):\penalty0 e0164828, 2016.

\bibitem[Anderson \& Agarwal(2011)Anderson and
  Agarwal]{anderson2011digitization}
Anderson, C.~L. and Agarwal, R.
\newblock The digitization of healthcare: boundary risks, emotion, and consumer
  willingness to disclose personal health information.
\newblock \emph{Information Systems Research}, 22\penalty0 (3):\penalty0
  469--490, 2011.

\bibitem[Arellano et~al.(2018)Arellano, Dai, Wang, Jiang, and
  Ohno-Machado]{arellano2018privacy}
Arellano, A.~M., Dai, W., Wang, S., Jiang, X., and Ohno-Machado, L.
\newblock Privacy policy and technology in biomedical data science.
\newblock \emph{Annual review of biomedical data science}, 1:\penalty0 115,
  2018.

\bibitem[Atan et~al.(2016)Atan, Whoiles, and Schaar]{schaar_costly}
Atan, O., Whoiles, W., and Schaar, M.
\newblock Data-driven online decision making with costly information
  acquisition.
\newblock \emph{Arxiv}, 02 2016.

\bibitem[Auer et~al.(2008)Auer, Jaksch, and Ortner]{auer2008near}
Auer, P., Jaksch, T., and Ortner, R.
\newblock Near-optimal regret bounds for reinforcement learning.
\newblock \emph{Advances in neural information processing systems}, 21, 2008.

\bibitem[Auxier et~al.(2019)Auxier, Rainie, Anderson, Perrin, Kumar, and
  Turner]{auxier2019americans}
Auxier, B., Rainie, L., Anderson, M., Perrin, A., Kumar, M., and Turner, E.
\newblock Americans and privacy: Concerned, confused and feeling lack of
  control over their personal information.
\newblock \emph{Pew Research Center: Internet, Science and Tech}, 2019.

\bibitem[Awad \& Krishnan(2006)Awad and Krishnan]{awad2006personalization}
Awad, N.~F. and Krishnan, M.~S.
\newblock The personalization privacy paradox: an empirical evaluation of
  information transparency and the willingness to be profiled online for
  personalization.
\newblock \emph{MIS quarterly}, pp.\  13--28, 2006.

\bibitem[Bansal et~al.(2010)Bansal, Gefen, et~al.]{bansal2010impact}
Bansal, G., Gefen, D., et~al.
\newblock The impact of personal dispositions on information sensitivity,
  privacy concern and trust in disclosing health information online.
\newblock \emph{Decision support systems}, 49\penalty0 (2):\penalty0 138--150,
  2010.

\bibitem[Bertsimas \& Kallus(2020)Bertsimas and
  Kallus]{bertsimas2020predictive}
Bertsimas, D. and Kallus, N.
\newblock From predictive to prescriptive analytics.
\newblock \emph{Management Science}, 66\penalty0 (3):\penalty0 1025--1044,
  2020.

\bibitem[Bertsimas et~al.(2019)Bertsimas, Dunn, and
  Mundru]{bertsimas2019optimal}
Bertsimas, D., Dunn, J., and Mundru, N.
\newblock Optimal prescriptive trees.
\newblock \emph{INFORMS Journal on Optimization}, 1\penalty0 (2):\penalty0
  164--183, 2019.

\bibitem[Biega et~al.(2020)Biega, Potash, Daum{\'e}, Diaz, and
  Finck]{biega2020operationalizing}
Biega, A.~J., Potash, P., Daum{\'e}, H., Diaz, F., and Finck, M.
\newblock Operationalizing the legal principle of data minimization for
  personalization.
\newblock In \emph{Proceedings of the 43rd international ACM SIGIR conference
  on research and development in information retrieval}, pp.\  399--408, 2020.

\bibitem[Bien et~al.(2013)Bien, Taylor, and Tibshirani]{bien2013lasso}
Bien, J., Taylor, J., and Tibshirani, R.
\newblock A lasso for hierarchical interactions.
\newblock \emph{Annals of statistics}, 41\penalty0 (3):\penalty0 1111, 2013.

\bibitem[Biggs et~al.(2020)Biggs, Sun, and Ettl]{biggs2020model}
Biggs, M., Sun, W., and Ettl, M.
\newblock Model distillation for revenue optimization: Interpretable
  personalized pricing.
\newblock \emph{arXiv preprint arXiv:2007.01903}, 2020.

\bibitem[Bukaty(2019)]{ccpa}
Bukaty, P.
\newblock \emph{The California Consumer Privacy Act (CCPA): An implementation
  guide}.
\newblock IT Governance Publishing, 2019.
\newblock ISBN 9781787781337.
\newblock URL \url{https://books.google.com/books?id=vGWfDwAAQBAJ}.

\bibitem[Campbell \& Kracaw(1980)Campbell and Kracaw]{campbell1980information}
Campbell, T.~S. and Kracaw, W.~A.
\newblock Information production, market signalling, and the theory of
  financial intermediation.
\newblock \emph{the Journal of Finance}, 35\penalty0 (4):\penalty0 863--882,
  1980.

\bibitem[Chemmanur(1993)]{chemmanur1993pricing}
Chemmanur, T.~J.
\newblock The pricing of initial public offerings: A dynamic model with
  information production.
\newblock \emph{The Journal of Finance}, 48\penalty0 (1):\penalty0 285--304,
  1993.

\bibitem[DeLong et~al.(1988)DeLong, DeLong, and
  Clarke-Pearson]{delong1988comparing}
DeLong, E.~R., DeLong, D.~M., and Clarke-Pearson, D.~L.
\newblock Comparing the areas under two or more correlated receiver operating
  characteristic curves: a nonparametric approach.
\newblock \emph{Biometrics}, pp.\  837--845, 1988.

\bibitem[DiCiccio \& Efron(1996)DiCiccio and Efron]{diciccio1996bootstrap}
DiCiccio, T.~J. and Efron, B.
\newblock Bootstrap confidence intervals.
\newblock \emph{Statistical science}, pp.\  189--212, 1996.

\bibitem[Dietterich(1998)]{dietterich1998approximate}
Dietterich, T.~G.
\newblock Approximate statistical tests for comparing supervised classification
  learning algorithms.
\newblock \emph{Neural computation}, 10\penalty0 (7):\penalty0 1895--1923,
  1998.

\bibitem[Do et~al.(2021)Do, Corbett-Davies, Atif, and Usunier]{do2021online}
Do, V., Corbett-Davies, S., Atif, J., and Usunier, N.
\newblock Online certification of preference-based fairness for personalized
  recommender systems.
\newblock \emph{arXiv preprint arXiv:2104.14527}, 2021.

\bibitem[Edwards et~al.(2013)Edwards, Naik, Ahmed, Elwyn, Pickles, Hood, and
  Playle]{edwards2013personalised}
Edwards, A.~G., Naik, G., Ahmed, H., Elwyn, G.~J., Pickles, T., Hood, K., and
  Playle, R.
\newblock Personalised risk communication for informed decision making about
  taking screening tests.
\newblock \emph{Cochrane database of systematic reviews}, Cochrane database of
  systematic reviews\penalty0 (2), 2013.

\bibitem[Elmachtoub et~al.(2018)Elmachtoub, Gupta, and
  Hamilton]{elmachtoub2018value}
Elmachtoub, A.~N., Gupta, V., and Hamilton, M.
\newblock The value of personalized pricing.
\newblock \emph{Available at SSRN 3127719}, 2018.

\bibitem[Estrada et~al.(1999)Estrada, Barnes, Collins, and
  Byrd]{estrada1999health}
Estrada, C., Barnes, V., Collins, C., and Byrd, J.~C.
\newblock Health literacy and numeracy.
\newblock \emph{Jama}, 282\penalty0 (6):\penalty0 527--527, 1999.

\bibitem[{European Parliament and of the Council}(2016)]{gdpr}
{European Parliament and of the Council}.
\newblock Regulation 2016/679 of the european parliament and of the council of
  27 april 2016 on the protection of natural persons with regard to the
  processing of personal data and on the free movement of such data, and
  repealing directive 95/46/ec (general data protection regulation), 2016.
\newblock URL \url{https://eur-lex.europa.eu/eli/reg/2016/679/oj}.
\newblock Official Journal of the European Union.

\bibitem[Fan \& Poole(2006)Fan and Poole]{fan2006personalization}
Fan, H. and Poole, M.~S.
\newblock What is personalization? perspectives on the design and
  implementation of personalization in information systems.
\newblock \emph{Journal of Organizational Computing and Electronic Commerce},
  16\penalty0 (3-4):\penalty0 179--202, 2006.

\bibitem[Globus-Harris et~al.(2022)Globus-Harris, Kearns, and
  Roth]{2022biasbounties}
Globus-Harris, I., Kearns, M., and Roth, A.
\newblock An algorithmic framework for bias bounties.
\newblock \emph{2022 ACM Conference on Fairness, Accountability, and
  Transparency}, Jun 2022.
\newblock \doi{10.1145/3531146.3533172}.
\newblock URL \url{http://dx.doi.org/10.1145/3531146.3533172}.

\bibitem[Haukoos et~al.(2012)Haukoos, Lyons, Lindsell, Hopkins, Bender,
  Rothman, Hsieh, MacLaren, Thrun, Sasson, et~al.]{denverhiv}
Haukoos, J.~S., Lyons, M.~S., Lindsell, C.~J., Hopkins, E., Bender, B.,
  Rothman, R.~E., Hsieh, Y.-H., MacLaren, L.~A., Thrun, M.~W., Sasson, C.,
  et~al.
\newblock Derivation and validation of the denver human immunodeficiency virus
  (hiv) risk score for targeted hiv screening.
\newblock \emph{American journal of epidemiology}, 175\penalty0 (8):\penalty0
  838--846, 2012.

\bibitem[H{\'e}bert-Johnson et~al.(2018)H{\'e}bert-Johnson, Kim, Reingold, and
  Rothblum]{hebert2018multicalibration}
H{\'e}bert-Johnson, {\'U}., Kim, M., Reingold, O., and Rothblum, G.
\newblock Multicalibration: Calibration for the (computationally-identifiable)
  masses.
\newblock In \emph{Proceedings of the International Conference on Machine
  Learning}, pp.\  1944--1953, 2018.

\bibitem[Hollenberg(2003)]{hollenberg2003cardiogenic}
Hollenberg, S.
\newblock Cardiogenic shock.
\newblock In \emph{Intensive Care Medicine}, pp.\  447--458. Springer, 2003.

\bibitem[Johnson et~al.(2016)Johnson, Pollard, Shen, Li-Wei, Feng, Ghassemi,
  Moody, Szolovits, Celi, and Mark]{johnson2016mimic}
Johnson, A.~E., Pollard, T.~J., Shen, L., Li-Wei, H.~L., Feng, M., Ghassemi,
  M., Moody, B., Szolovits, P., Celi, L.~A., and Mark, R.~G.
\newblock Mimic-iii, a freely accessible critical care database.
\newblock \emph{Scientific data}, 3\penalty0 (1):\penalty0 1--9, 2016.

\bibitem[Kaminski(2019)]{kaminski2019right}
Kaminski, M.~E.
\newblock The right to explanation, explained.
\newblock \emph{Berkeley Tech. LJ}, 34:\penalty0 189, 2019.

\bibitem[Kearns et~al.(2018)Kearns, Neel, Roth, and Wu]{kearns2018preventing}
Kearns, M., Neel, S., Roth, A., and Wu, Z.~S.
\newblock Preventing fairness gerrymandering: Auditing and learning for
  subgroup fairness.
\newblock In \emph{International Conference on Machine Learning}, pp.\
  2564--2572, 2018.

\bibitem[Kessler et~al.(2005)Kessler, Adler, Ames, Demler, Faraone, Hiripi,
  Howes, Jin, Secnik, Spencer, et~al.]{kessler2005world}
Kessler, R.~C., Adler, L., Ames, M., Demler, O., Faraone, S., Hiripi, E.,
  Howes, M.~J., Jin, R., Secnik, K., Spencer, T., et~al.
\newblock The world health organization adult adhd self-report scale (asrs): a
  short screening scale for use in the general population.
\newblock \emph{Psychological medicine}, 35\penalty0 (2):\penalty0 245--256,
  2005.

\bibitem[Kim et~al.(2019)Kim, Korolova, Rothblum, and Yona]{kim2019preference}
Kim, M.~P., Korolova, A., Rothblum, G.~N., and Yona, G.
\newblock Preference-informed fairness.
\newblock \emph{arXiv preprint arXiv:1904.01793}, 2019.

\bibitem[Le~Gall et~al.(1993)Le~Gall, Lemeshow, and Saulnier]{le1993new}
Le~Gall, J.-R., Lemeshow, S., and Saulnier, F.
\newblock A new simplified acute physiology score (saps ii) based on a
  european/north american multicenter study.
\newblock \emph{Jama}, 270\penalty0 (24):\penalty0 2957--2963, 1993.

\bibitem[Lim \& Hastie(2015)Lim and Hastie]{lim2015learning}
Lim, M. and Hastie, T.
\newblock Learning interactions via hierarchical group-lasso regularization.
\newblock \emph{Journal of Computational and Graphical Statistics}, 24\penalty0
  (3):\penalty0 627--654, 2015.

\bibitem[Lundberg et~al.(2019)Lundberg, Narayanan, Levy, and
  Salganik]{lundberg2019privacy}
Lundberg, I., Narayanan, A., Levy, K., and Salganik, M.~J.
\newblock Privacy, ethics, and data access: A case study of the fragile
  families challenge.
\newblock \emph{Socius}, 5:\penalty0 2378023118813023, 2019.

\bibitem[McAuley(in press)]{mcauley2022pml}
McAuley, J.
\newblock \emph{Personalized Machine Learning}.
\newblock Cambridge University Press, in press.

\bibitem[Nanayakkara et~al.(2022)Nanayakkara, Bater, He, Hullman, and
  Rogers]{nanayakkara2022visualizing}
Nanayakkara, P., Bater, J., He, X., Hullman, J., and Rogers, J.
\newblock Visualizing privacy-utility trade-offs in differentially private data
  releases.
\newblock \emph{arXiv preprint arXiv:2201.05964}, 2022.

\bibitem[OECD(2013)]{oecd2013}
OECD.
\newblock Recommendation of the council concerning guidelines governing the
  protection of privacy and transborder flows of personal data, 2013.
\newblock URL
  \url{https://legalinstruments.oecd.org/en/instruments/OECD-LEGAL-0188}.

\bibitem[Ortlieb \& Garner(2016)Ortlieb and Garner]{ortlieb2016sensitivity}
Ortlieb, M. and Garner, R.
\newblock Sensitivity of personal data items in different online contexts.
\newblock \emph{it-Information Technology}, 58\penalty0 (5):\penalty0 217--228,
  2016.

\bibitem[Paes et~al.(2022)Paes, Long, Ustun, and Calmon]{paes2022on}
Paes, L.~M., Long, C.~X., Ustun, B., and Calmon, F.
\newblock On the epistemic limits of personalized prediction.
\newblock In Oh, A.~H., Agarwal, A., Belgrave, D., and Cho, K. (eds.),
  \emph{Advances in Neural Information Processing Systems}, 2022.
\newblock URL \url{https://openreview.net/forum?id=Snp3iEj7NJ}.

\bibitem[Paulus \& Kent(2020)Paulus and Kent]{paulus2020predictably}
Paulus, J.~K. and Kent, D.~M.
\newblock Predictably unequal: understanding and addressing concerns that
  algorithmic clinical prediction may increase health disparities.
\newblock \emph{NPJ digital medicine}, 3\penalty0 (1):\penalty0 1--8, 2020.

\bibitem[Pollard et~al.(2018)Pollard, Johnson, Raffa, Celi, Mark, and
  Badawi]{pollard2018eicu}
Pollard, T.~J., Johnson, A.~E., Raffa, J.~D., Celi, L.~A., Mark, R.~G., and
  Badawi, O.
\newblock The eicu collaborative research database, a freely available
  multi-center database for critical care research.
\newblock \emph{Scientific data}, 5\penalty0 (1):\penalty0 1--13, 2018.

\bibitem[Scosyrev et~al.(2012)Scosyrev, Messing, Noyes, Veazie, and
  Messing]{nci2019seer}
Scosyrev, E., Messing, J., Noyes, K., Veazie, P., and Messing, E.
\newblock Surveillance epidemiology and end results (seer) program and
  population-based research in urologic oncology: an overview.
\newblock In \emph{Urologic Oncology: Seminars and Original Investigations},
  volume~30, pp.\  126--132. Elsevier, 2012.

\bibitem[Shanmugam et~al.(2022)Shanmugam, Diaz, Shabanian, Finck, and
  Biega]{shanmugam2022learning}
Shanmugam, D., Diaz, F., Shabanian, S., Finck, M., and Biega, A.
\newblock Learning to limit data collection via scaling laws: A computational
  interpretation for the legal principle of data minimization.
\newblock In \emph{Proceedings of the 2022 ACM Conference on Fairness,
  Accountability, and Transparency}, pp.\  839--849, 2022.

\bibitem[Sharp(2015)]{sharp2015hamilton}
Sharp, R.
\newblock The hamilton rating scale for depression.
\newblock \emph{Occupational Medicine}, 65\penalty0 (4):\penalty0 340--340,
  2015.

\bibitem[Steyerberg et~al.(2019)]{steyerberg2019clinical}
Steyerberg, E.~W. et~al.
\newblock \emph{Clinical prediction models}.
\newblock Springer, 2019.

\bibitem[Struck et~al.(2017)Struck, Ustun, Ruiz, Lee, LaRoche, Hirsch, Gilmore,
  Vlachy, Haider, and Rudin]{struck2017association}
Struck, A.~F., Ustun, B., Ruiz, A.~R., Lee, J.~W., LaRoche, S.~M., Hirsch,
  L.~J., Gilmore, E.~J., Vlachy, J., Haider, H.~A., and Rudin, C.
\newblock Association of an electroencephalography-based risk score with
  seizure probability in hospitalized patients.
\newblock \emph{JAMA neurology}, 74\penalty0 (12):\penalty0 1419--1424, 2017.

\bibitem[Struck et~al.(2020)Struck, Tabaeizadeh, Schmitt, Ruiz, Swisher,
  Subramaniam, Hernandez, Kaleem, Haider, and Ciss{\'e}]{struck2020assessment}
Struck, A.~F., Tabaeizadeh, M., Schmitt, S.~E., Ruiz, A.~R., Swisher, C.~B.,
  Subramaniam, T., Hernandez, C., Kaleem, S., Haider, H.~A., and Ciss{\'e},
  A.~F.
\newblock Assessment of the validity of the 2helps2b score for inpatient
  seizure risk prediction.
\newblock \emph{JAMA neurology}, 77\penalty0 (4):\penalty0 500--507, 2020.

\bibitem[Sun \& Xu(2014)Sun and Xu]{fastdelong}
Sun, X. and Xu, W.
\newblock Fast implementation of delong's algorithm for comparing the areas
  under correlated receiver operating characteristic curves.
\newblock \emph{IEEE Signal Processing Letters}, 21\penalty0 (11):\penalty0
  1389--1393, 2014.
\newblock \doi{10.1109/LSP.2014.2337313}.

\bibitem[Suriyakumar et~al.(2023)Suriyakumar, Ghassemi, and
  Ustun]{suriyakumar2022fairuse}
Suriyakumar, V.~M., Ghassemi, M., and Ustun, B.
\newblock When personalization harms: Reconsidering the use of group attributes
  in prediction.
\newblock In \emph{International Conference on Machine Learning}, 2023.

\bibitem[Tran \& Fioretto(2023)Tran and Fioretto]{tran2023personalized}
Tran, C. and Fioretto, F.
\newblock Personalized privacy auditing and optimization at test time.
\newblock \emph{arXiv preprint arXiv:2302.00077}, 2023.

\bibitem[Tran et~al.(2021)Tran, Dinh, and Fioretto]{tran2021differentially}
Tran, C., Dinh, M., and Fioretto, F.
\newblock Differentially private empirical risk minimization under the fairness
  lens.
\newblock \emph{Advances in Neural Information Processing Systems},
  34:\penalty0 27555--27565, 2021.

\bibitem[{U.S. Congress}(1996)]{hipaa}
{U.S. Congress}.
\newblock Health insurance portability and accountability act of 1996, 1996.
\newblock URL
  \url{https://www.hhs.gov/hipaa/for-professionals/privacy/index.html}.
\newblock Public Law 104-191.

\bibitem[Ustun et~al.(2016)Ustun, Westover, Rudin, and
  Bianchi]{ustun2016clinical}
Ustun, B., Westover, M.~B., Rudin, C., and Bianchi, M.~T.
\newblock Clinical prediction models for sleep apnea: the importance of medical
  history over symptoms.
\newblock \emph{Journal of Clinical Sleep Medicine}, 12\penalty0 (02):\penalty0
  161--168, 2016.

\bibitem[Ustun et~al.(2019)Ustun, Liu, and Parkes]{ustun2019fairness}
Ustun, B., Liu, Y., and Parkes, D.
\newblock Fairness without harm: Decoupled classifiers with preference
  guarantees.
\newblock In \emph{International Conference on Machine Learning}, pp.\
  6373--6382, 2019.

\bibitem[Vaughan et~al.(2020)Vaughan, Aseltine, Chen, and
  Yan]{vaughan2020efficient}
Vaughan, G., Aseltine, R., Chen, K., and Yan, J.
\newblock Efficient interaction selection for clustered data via stagewise
  generalized estimating equations.
\newblock \emph{Statistics in Medicine}, 39\penalty0 (22):\penalty0 2855--2868,
  2020.

\bibitem[Viviano \& Bradic(2020)Viviano and Bradic]{viviano2020fair}
Viviano, D. and Bradic, J.
\newblock Fair policy targeting.
\newblock \emph{arXiv preprint arXiv:2005.12395}, 2020.

\bibitem[Vyas et~al.(2020)Vyas, Eisenstein, and Jones]{vyas2020hidden}
Vyas, D.~A., Eisenstein, L.~G., and Jones, D.~S.
\newblock Hidden in plain sight---reconsidering the use of race correction in
  clinical algorithms, 2020.

\bibitem[Yu et~al.(2009)Yu, Krishnapuram, Rosales, and Rao]{Yu2009ActiveS}
Yu, S., Krishnapuram, B., Rosales, R., and Rao, R.~B.
\newblock Active sensing.
\newblock In \emph{AISTATS}, 2009.

\bibitem[Zafar et~al.(2017)Zafar, Valera, Rodriguez, Gummadi, and
  Weller]{zafar2017parity}
Zafar, M.~B., Valera, I., Rodriguez, M., Gummadi, K., and Weller, A.
\newblock From parity to preference-based notions of fairness in
  classification.
\newblock In \emph{Advances in Neural Information Processing Systems}, pp.\
  228--238, 2017.

\end{thebibliography}

\appendix
\renewcommand\thepart{}
\renewcommand\partname{}
\mtcsetrules{parttoc}{off}
\mtcsetoffset{parttoc}{-2.5em}
\def\mtcgapafterheads{20em}
\part{\texorpdfstring{}{Appendices}} 
\let\oldsection\section
\renewcommand\section{\clearpage\oldsection}

\iftoggle{icml}{
\begin{center}
\hrule height 4pt
\vskip 0.25in
\vskip -\parskip
{\LARGE\bfseries\ourtitle{}}
\vskip 0.29in
\vskip -\parskip
\hrule height 1pt
\vskip 0.09in

{\Large\bfseries Supplementary Material}
\end{center}
\renewcommand\ptctitle{}
\parttoc 
\setcounter{page}{0}
\thispagestyle{empty}
}%

\iftoggle{neurips}{
\clearpage
\begin{center}
%

{\Large\bfseries Supplementary Material}
\end{center}
\renewcommand\ptctitle{}
\parttoc 
\setcounter{page}{0}
\thispagestyle{empty}
}%

\iftoggle{arxiv}{
\renewcommand\ptctitle{Appendices}
\parttoc 
}

\clearpage
\section{Supporting Material for \cref{Sec::Algorithms}}
\label{Appendix::Algorithms}

%
\newcommand{\A}{\mathcal{A}}
\newcommand{\U}{\mathcal{U}}
\renewcommand{\P}{\Pi}
\renewcommand{\trees}{\alltrees{}}
\newcommand{\allsubtrees}{\mathcal{S}}

\subsection{Enumeration Routine for \cref{Alg::ParticipatorySystem}}

We summarize the Enumeration routine in \cref{Alg::EnumerateTrees}. \cref{Alg::EnumerateTrees} takes as input a set of group attributes $\G$ and a dataset $\data{}$ and outputs a collection of reporting interfaces $\trees{}$ that obey ordering and plausibility constraints. 
\begin{algorithm*}[h]
\begin{algorithmic}[1]
\small
\Procedure{\textsf{ViableTrees}}{$\G, \data{}$}
\State \textbf{if} {dim($\G{}$) = 1} \textbf{return} [$T_\G{}$] \Comment{base case: we are left with only a single attribute on which to branch}
\State $\trees \gets [\ ]$
\For{each group attribute $\A \in [\G_1,\ldots,\G_k]$}\label{AlgStep::EnumerateFor}
    \State $T_\A{} \gets$ reporting tree of depth 1 with $|\A|$ leaves
    \State $\allsubtrees \gets \textsf{ViableTrees}(\G \setminus \A, \data{})$ \Comment{all subtrees using all attributes except $\A$}
    \For{$\P{}$ in \textsf{ValidAssignments}($\allsubtrees{},\A, \data{}$)}: \Comment{each assignment is a permutation of $|\A|$ to leaves of $T_\A{}$}
        \State $\trees{} \gets \trees{} \cup T_{\A}.\textsf{assign}(\P{})$ \Comment{extends the tree by assigning subtrees to each leaf}
    \EndFor
\EndFor
\State \Return{$\trees{}$, reporting interfaces for group attributes $\G{}$ that obey plausibility and ordering constraints}
\EndProcedure
\end{algorithmic}
\caption{Enumerate All Possible Reporting Trees for Reporting Options $\G{}$}
\label{Alg::EnumerateTrees}
\end{algorithm*}

The routine enumerates all possible reporting interfaces for a given set of group attributes $\G$ through a recursive branching process. Given a set of group attributes, the routine is called for each attribute that has yet to be considered in the tree \cref{AlgStep::EnumerateFor}, ensuring a complete enumeration. We note that the routine is only called for building Sequential systems since there is only one possible reporting interface for Minimal and Flat systems.

Enumerating all possible trees ensures we can recover the best tree given the selection criteria and allows practitioners to choose between models based on other criteria. We generate trees that meet plausibility constraints based on the dataset, such as having at least one negative and one positive sample and at least $s$ total samples at each leaf.  In settings constrained by computational resources, we can impose additional stopping criteria and modify the ordering to enumerate more plausible trees first or exclusively (e.g., by changing the ordering of $\G$ or imposing constraints in \textsc{ValidAssignments}). 

\FloatBarrier
\subsection{Assignment Routine for \cref{Alg::ParticipatorySystem}}

We summarize the routine for \textprc{AssignModels} procedure in \cref{Alg::Assign}. 
\begin{algorithm*}[!h]
\begin{algorithmic}[1]
\small
\Procedure{\textsf{AssignModels}}{$\tree, \candidatepool{}, \data{}$}
\State $Q \gets [\tree.\textsf{root}]$ \Comment{initialize with the root of the tree, reporting group $\Dnr{}$}
\While{$Q$ is not empty}
\State $\rg \gets Q.\textsf{pop}()$
\State $\candidatepool_{\rg} \gets \textsf{ViableModels}(\candidatepool, \rg)$ \Comment{filter $\candidatepool$ to models that can be assigned to $\rg$}
\State $\displaystyle h^* \gets \argmin_{h \in \candidatepool_{\rg}}{\emprisk{\rg}{h, \data{}}}$\Comment{assign the model with the best training performance}
\State $\tree.\textsf{set\_model}(\rg, h^*)$
\For{$\rg' \in \tree.\textsf{get\_subgroups}(\rg)$} \Comment{iterate through the children reporting groups of $\rg$}
\State $Q.\textsf{enqueue}(\rg')$
\EndFor
\EndWhile
\State \Return{$\tree$ that maximizes gain for each reporting group}
\EndProcedure
\end{algorithmic}
\caption{Assigning Models}
\label{Alg::Assign}
\end{algorithm*}

\cref{Alg::Assign} takes as inputs a reporting tree $\tree{}$, a pool candidate models $\candidatepool$, and an assignment (training) dataset $\data{}$ and outputs a tree $\tree$ that maximizes the gains of reporting group information. The pool of candidate models is filtered to viable models for each reporting group. Since the pool of candidate models includes the generic model $\clf{}$, each reporting group will have at least one viable model. We assign each reporting group the best-performing model on the training set and default to the generic model $\clf{}$ when a better-performing personalized model is not found. We assign performance on the training set and then prune using performance on the validation set to avoid biased gain estimations.

\FloatBarrier
\subsection{Pruning Routine for \cref{Alg::ParticipatorySystem}}

We summarize the routine used for the \textprc{PruneLeaves} procedure in \cref{Alg::ParticipatorySystem}. The \textprc{PruneLeaves} routine 

\newcommand{\leafstack}[1]{L}
\begin{algorithm*}[!h]
\begin{algorithmic}[1]
\small
\Procedure{\textsf{PruneLeaves}}{$\tree, \data{}$}
\State $Stack \gets [\tree.\textsf{leaves}]$ \Comment{initialize stack with all leaves}
\Repeat{}
\State $\rg \gets Stack.pop()$
\State $\plf{} \gets \tree.\textsf{get\_model}(\rg)$
\State $\plf{}' \gets \tree.\textsf{get\_model}(\parent{\rg})$
\If{not $\textprc{Test}(\rg, \plf{},\plf{}', \data{})$}\Comment{test gains to see if parent model is as good as leaf model}
\State $\tree.\textsf{prune}(\rg)$ 
\EndIf
\If{$\tree.\textsf{get\_children}(\parent{\rg})$ is empty} \Comment{consider pruning the parent if the parent has become a leaf}
\State $Stack.\textsf{enqueue}(\parent{\rg})$
\EndIf
\Until{$Stack$ is empty}
\State \Return{$\tree$, reporting interface that ensures data collection leads to gain}
\EndProcedure
\end{algorithmic}
\caption{Pruning Participatory Systems}
\label{Alg::Prune}
\end{algorithm*}

\cref{Alg::ParticipatorySystem} takes as input a reporting interface $\tree$ and a validation sample $\data{}$, and performs a bottom-up pruning to output a reporting interface $\tree$ that asks individuals to report attributes that are expected to lead to a gain.
The pruning decision at each leaf is based on a hypothesis test that evaluates the gains of reporting for a reporting group on a validation dataset. This test has the form:
\begin{align*}
H_0: \truerisk{\gb}{h} \leq \truerisk{\gb}{h'} \quad\text{vs.}\quad H_A: \truerisk{\gb}{h} > \truerisk{\gb}{h'}
\end{align*}
This procedure evaluates the gains of reporting by comparing the performance of a model assigned at a leaf node $h$ and a model assigned at a parent node $h'$ which does not use the reported information. 
%
%
Here, the null hypothesis $H_0$ assumes that the parent model performs as well as the leaf model -- and thus, we reject the null hypothesis when there is sufficient evidence to suggest that reporting will improve performance in deployment. Our routine allows practitioners to specify the hypothesis test to compute the gains. By default, we use the McNemar test for accuracy \cite{dietterich1998approximate} and the Delong test for AUC \cite{delong1988comparing, fastdelong}. In general, we can use a bootstrap hypothesis test~\citep[][]{diciccio1996bootstrap}. 

\FloatBarrier
\subsection{Greedy Induction of Sequential Reporting Interface}
\label{Appendix::GreedyAlgorithm}

We present an additional routine to construct reporting interfaces for sequential systems in \cref{Alg::GreedyTree}. We include this routine as an alternative option that can be used to construct a reporting interface in settings where it may be impractical or undesirable to enumerate all possible reporting interfaces. The procedure results in a valid reporting interface that ensures gains. However, it does not guarantee an optimal tree in terms of maximizing the overall gain and does not allow to practitioners to choose between reporting interfaces after training. 

\begin{algorithm}[thb]
\begin{algorithmic}[1]
\small
\Procedure{\textsf{GreedyTree}}{$\RG$}
\State $\tree{} \gets$ empty reporting interface
\Repeat{}
\For{$\node \in \textrm{leaves}(\tree{})$}
\State $\{\A_{\node}\} \gets {G_i : \node[i] = \dnr}$
\Comment{$\{\A_{\node}\}$ contains all heretofore unused attributes}
\State $\A^* \gets \argmax_{\A \in \{\A_{\node}\}}{ \min_{\rg' \in \node.\textsf{split}(\A)}{\truegap{\rg'}{\rg'}{\rg}}}$
\State $\node.\textsf{split}(\A^{*})$
\Comment{Split on attribute that maximizes worse-case gain}
\EndFor{}
\Until{no splits are added}
\State \Return{$\tree{}$, reporting interface that ensures gains for reporting each $\RG{}$.}
\EndProcedure
\end{algorithmic}
\caption{Greedy Induction Routine for Sequential Reporting Interfaces}
\label{Alg::GreedyTree}
\end{algorithm}
\cref{Alg::GreedyTree} takes as input a collection of reporting options $\RG$ and outputs a single reporting interface using a greedy tree induction routine that chooses the attribute to report to maximize the minimum gain at each step. The procedure uses the reporting options to iteratively construct a reporting tree that branches on all of the attributes in $\RG$. The procedure considers each unused attribute for each splitting point and splits on the attribute that provides the greatest minimum gain for the groups contained at that node. 

\clearpage
\section{Description of Datasets used in \cref{Sec::Experiments} -- Experiments}
\label{Appendix::Datasets}

We include additional information about the datasets used in \cref{Sec::Experiments}. 

\begin{table}[ht]
\centering
\resizebox{\linewidth}{!}{
\begin{tabular}{lllrrrc}
\toprule
\cell{l}{\textbf{Dataset}} &  
\cell{l}{\textbf{Reference}} & 
\cell{l}{\textbf{Outcome Variable}} &
\cell{r}{\large{$n$}} & 
\cell{r}{\large{$d$}} & 
\cell{r}{\large{$m$}} & 
$\G$ \\
\toprule



\textds{apnea} & 
\citet{ustun2016clinical} &
patient has obstructive sleep apnea & 
1,152 & 28 & 6 & 
\{\textgp{age}, \textgp{sex}\}\\

\midrule

\textds{cardio\_eicu} & 
\citet{pollard2018eicu} &
patient with cardiogenic shock dies & 
1,341 & 49 & 8 & 
\{\textgp{age}, \textgp{sex}, \textgp{race}\}\\
\midrule

\textds{cardio\_mimic} & 
\citet{johnson2016mimic} &
patient with cardiogenic shock dies & 
5,289 & 49 & 8 & 
\{\textgp{age}, \textgp{sex}, \textgp{race}\}\\
\midrule

\textds{coloncancer} & 
\citet[][]{nci2019seer} &
patient dies within 5 years & 29,211 & 72 & 6 & 
\{\textgp{age}, \textgp{sex}\}\\
\midrule

\textds{lungcancer} & 
\citet[][]{nci2019seer} &
patient dies within 5 years & 
120,641 & 84 & 6 & 
\{\textgp{age}, \textgp{sex}\}\\

\midrule
\textds{saps} & 
\citet{allyn2016simplified} &
ICU mortality & 
7,797 & 36 & 4 & 
\{\textgp{age}, \textgp{HIV}\}\\

\bottomrule
\end{tabular}
}
\caption{Datasets used to fit clinical prediction models in \cref{Sec::Experiments}. Here: $n$ denotes the number of examples in each dataset; $d$ denotes the number of features; $\G$ denotes the group attributes that are used for personalization; and $m = |\G|$ denotes the number of intersectional groups. Each dataset is de-identified and available to the public. The \textds{cardio\_eicu}, \textds{cardio\_mimic}, \textds{lungcancer} datasets require access to public repositories listed under the references. The \textds{saps} and \textds{apnea} datasets must be requested from the authors. The \textds{support} dataset can be downloaded directly from the URL below.}
\label{Table::Datasets}
\end{table}

\paragraph{\textds{apnea}} We use the obstructive sleep apnea (OSA) dataset outlined in~\citet{ustun2016clinical}. This dataset includes a cohort of 1,152 patients where 23\% have OSA. We use all available features (e.g. BMI, comorbidities, age, and sex) and binarize them, resulting in 26 binary features.

\paragraph{\textds{cardio\_eicu} \& \textds{cardio\_mimic}} Cardiogenic shock is an acute condition in which the heart cannot provide sufficient blood to the vital organs~\citep{hollenberg2003cardiogenic}. These datasets are designed to predict cardiogenic shock for patients in intensive care. Each dataset contains the same features, group attributes, and outcome variables for patients in different cohorts. The \textds{cardio\_eicu} dataset contains records for a cohort of patients in the Collaborative Research Database V2.0 \citep{pollard2018eicu}. The \textds{cardio\_eicu} dataset contains records for a cohort of patients in the MIMIC-III~\citep{johnson2016mimic} database. Here, the outcome variable indicates whether a patient in the ICU with cardiogenic shock will die while in the ICU. The features encode the results of vital signs and routine lab tests (e.g. systolic BP, heart rate, hemoglobin count) that were collected up to 24 hours before the onset of cardiogenic shock.

\paragraph{\textds{lungcancer}} We consider a cohort of 120,641 patients who were diagnosed with lung cancer between 2004-2016 and monitored as part of the National Cancer Institute SEER study \citep[][]{nci2019seer}. Here, the outcome variable indicates if a patient dies within five years from any cause, and 16.9\% of patients died within the first five years from diagnosis. The cohort includes patients from Greater California, Georgia, Kentucky, New Jersey, and Louisiana, and does not cover patients who were lost to follow-up (censored). Age and Sex were considered as group attributes. The features reflect the morphology and histology of the tumor (e.g., size, metastasis, stage,  node count and location, number and location of notes) as well as  interventions that were administered at the time of diagnosis (e.g., surgery, chemo, radiology).

\paragraph{\textds{coloncancer}} We consider a cohort of 120,641 patients who were diagnosed with colorectal cancer between 2004-2016 and monitored as part of the National Cancer Institute SEER study \citep[][]{nci2019seer}. Here, the outcome variable indicates if a patient dies within five years from any cause, and 42.1\% of patients die within the first five years from diagnosis. The cohort includes patients from Greater California. Age and Sex were considered as group attributes. The features reflect the morphology and histology of the tumor (e.g., size, metastasis, stage,  node count and location, number and location of notes) as well as  interventions that were administered at the time of diagnosis (e.g., surgery, chemo, radiology).

\paragraph{\textds{saps}} The Simplified Acute Physiology Score II (SAPS II) score predicts the risk of mortality of critically-ill patients in intensive care~\citep{le1993new}. The data contains records of 7,797 patients from 137 medical centers in 12 countries. Here, the outcome variable indicates whether a patient dies in the ICU, with 12.8\% patient of patients dying. The features reflect comorbidities, vital signs, and lab measurements.

\clearpage
\section{Results for Different Model Classes and Prediction Tasks}
\label{Appendix::ExtraResults}


In this Appendix, we present experimental results for additional model classes and prediction tasks. We produce these results using the setup in \cref{Sec::ExperimentalSetup}, and summarize them in the same way as \cref{Table::Results}. We refer to them in our discussion in \cref{Sec::ExperimentalResults}.



\subsection{Logistic Regression for Ranking (AUC)}
\label{Appendix::LogRegAuc}
\begin{table*}[!h]
\centering
\centering\begingroup\fontsize{8}{10}\selectfont
\renewcommand{\metricsguide}[0]{\cell{r}{Overall Performance\\Overall Gain\\Group Gains\\Max Disparity\\Rat. Violations\\Imputation Risk\\Options Pruned\\Data Use}}
\resizebox{\linewidth}{!}{
\begin{tabular}{lrlllllll}
\multicolumn{2}{c}{ } & \multicolumn{2}{c}{\textsc{Static}} & \multicolumn{2}{c}{\textsc{Imputed}} & \multicolumn{3}{c}{\textsc{Participatory}} \\
\cmidrule(l{3pt}r{3pt}){3-4} \cmidrule(l{3pt}r{3pt}){5-6} \cmidrule(l{3pt}r{3pt}){7-9}
Dataset & Metrics & \OneHot{} & \IntHot{} & \OneHotKNNImpute{} & \IntHotKNNImpute{} & \MinimalSys{} & \FlatSys{} & \SeqSys{}\\
\midrule
\apnea & \metricsguide{} & \cell{r}{0.774\\-0.002\\-0.002 -- 0.002\\0.004\\\badvalue{2}\\-0.002\\0/6\\100.0\%} & \cell{r}{0.774\\-0.002\\-0.002 -- 0.003\\0.005\\\badvalue{2}\\-0.002\\0/6\\100.0\%} & \cell{r}{0.776\\0.000\\-0.002 -- 0.002\\0.004\\\badvalue{2}\\\\0/12\\0.0\%} & \cell{r}{0.776\\-0.000\\-0.002 -- 0.003\\0.005\\\badvalue{2}\\\\0/12\\0.0\%} & \cell{r}{0.776\\0.000\\0.000 -- 0.002\\0.002\\0\\\\5/7\\16.7\%} & \cell{r}{\cellcolor{fgood}\textbf{0.851}\\\cellcolor{fgood}\textbf{0.074}\\0.004 -- 0.115\\0.111\\0\\\\4/12\\100.0\%} & \cell{r}{\cellcolor{fgood}\textbf{0.851}\\\cellcolor{fgood}\textbf{0.074}\\0.004 -- 0.115\\0.111\\0\\\\4/12\\83.3\%}\\
\midrule

\cshockReicu & \metricsguide{} & \cell{r}{0.864\\0.002\\-0.005 -- 0.003\\0.009\\\badvalue{3}\\-0.005\\0/8\\100.0\%} & \cell{r}{0.863\\0.001\\-0.010 -- 0.010\\0.019\\\badvalue{3}\\-0.010\\0/8\\100.0\%} & \cell{r}{0.863\\0.000\\-0.005 -- 0.003\\0.009\\\badvalue{3}\\\\0/27\\0.0\%} & \cell{r}{0.862\\-0.001\\-0.010 -- 0.010\\0.019\\\badvalue{3}\\\\0/27\\0.0\%} & \cell{r}{0.865\\0.002\\0.000 -- 0.003\\0.003\\0\\\\6/9\\25.0\%} & \cell{r}{0.966\\0.103\\0.010 -- 0.180\\0.170\\0\\\\13/27\\100.0\%} & \cell{r}{\cellcolor{fgood}\textbf{0.966}\\\cellcolor{fgood}\textbf{0.103}\\0.010 -- 0.180\\0.170\\0\\\\11/27\\95.8\%}\\
\midrule

\cshockRmimic & \metricsguide{} & \cell{r}{0.881\\0.000\\-0.001 -- 0.001\\0.002\\\badvalue{3}\\-0.001\\0/8\\100.0\%} & \cell{r}{0.881\\0.000\\-0.001 -- 0.001\\0.002\\\badvalue{3}\\-0.001\\0/8\\100.0\%} & \cell{r}{0.882\\0.002\\-0.001 -- 0.001\\0.002\\\badvalue{3}\\\\0/27\\0.0\%} & \cell{r}{0.880\\-0.000\\-0.001 -- 0.001\\0.002\\\badvalue{3}\\\\0/27\\0.0\%} & \cell{r}{0.881\\0.000\\0.000 -- 0.001\\0.001\\0\\\\6/9\\25.0\%} & \cell{r}{\cellcolor{fgood}\textbf{0.914}\\\cellcolor{fgood}\textbf{0.034}\\0.008 -- 0.057\\0.049\\0\\\\9/27\\100.0\%} & \cell{r}{\cellcolor{fgood}\textbf{0.914}\\\cellcolor{fgood}\textbf{0.034}\\0.008 -- 0.057\\0.049\\0\\\\8/27\\91.7\%}\\
\midrule

\seercrcagesexcalifornia & \metricsguide{} & \cell{r}{0.685\\0.001\\-0.001 -- 0.002\\0.003\\\badvalue{3}\\-0.001\\0/6\\100.0\%} & \cell{r}{0.685\\0.002\\-0.001 -- 0.001\\0.002\\\badvalue{2}\\-0.002\\0/6\\100.0\%} & \cell{r}{0.683\\-0.000\\-0.001 -- 0.002\\0.003\\\badvalue{3}\\\\0/12\\0.0\%} & \cell{r}{0.683\\-0.000\\-0.001 -- 0.001\\0.002\\\badvalue{2}\\\\0/12\\0.0\%} & \cell{r}{0.685\\0.001\\0.000 -- 0.001\\0.001\\0\\\\5/7\\16.7\%} & \cell{r}{\cellcolor{fgood}\textbf{0.700}\\\cellcolor{fgood}\textbf{0.016}\\0.001 -- 0.021\\0.020\\0\\\\2/12\\100.0\%} & \cell{r}{0.700\\0.016\\0.001 -- 0.021\\0.020\\0\\\\5/12\\75.0\%}\\
\midrule

\seerrespacmcollapsedagesexall & \metricsguide{} & \cell{r}{0.855\\0.001\\-0.000 -- 0.000\\0.001\\\badvalue{2}\\-0.000\\0/6\\100.0\%} & \cell{r}{0.855\\0.001\\-0.000 -- 0.000\\0.001\\\badvalue{2}\\-0.000\\0/6\\100.0\%} & \cell{r}{0.852\\-0.002\\-0.000 -- 0.000\\0.001\\\badvalue{2}\\\\0/12\\0.0\%} & \cell{r}{0.854\\0.000\\-0.000 -- 0.000\\0.001\\\badvalue{2}\\\\0/12\\0.0\%} & \cell{r}{0.855\\0.001\\0.000 -- 0.000\\0.000\\\badvalue{1}\\\\4/7\\33.3\%} & \cell{r}{\cellcolor{fgood}\textbf{0.861}\\\cellcolor{fgood}\textbf{0.006}\\0.001 -- 0.012\\0.011\\0\\\\2/12\\100.0\%} & \cell{r}{0.861\\0.006\\0.001 -- 0.012\\0.011\\0\\\\2/12\\91.7\%}\\
\midrule

\saps & \metricsguide{} & \cell{r}{0.875\\0.010\\-0.000 -- 0.016\\0.017\\\badvalue{1}\\-0.000\\0/4\\100.0\%} & \cell{r}{0.877\\0.011\\-0.002 -- 0.019\\0.021\\\badvalue{1}\\-0.002\\0/4\\100.0\%} & \cell{r}{0.875\\0.010\\-0.000 -- 0.016\\0.017\\\badvalue{1}\\\\0/9\\0.0\%} & \cell{r}{0.857\\-0.008\\-0.002 -- 0.019\\0.021\\\badvalue{1}\\\\0/9\\0.0\%} & \cell{r}{0.875\\0.009\\0.000 -- 0.016\\0.016\\0\\\\1/5\\75.0\%} & \cell{r}{\cellcolor{fgood}\textbf{0.960}\\\cellcolor{fgood}\textbf{0.095}\\0.035 -- 0.141\\0.106\\0\\\\2/9\\100.0\%} & \cell{r}{0.960\\0.095\\0.035 -- 0.141\\0.106\\0\\\\3/9\\87.5\%}\\

\bottomrule
\end{tabular}}
\endgroup{}

\caption{Overview of performance, data use, and consent for all personalized models and systems on all datasets as measured by \textbf{test auc}. We show the performance of models and systems built using \textbf{logistic regression}.}
\end{table*}

\clearpage
\subsection{Random Forests for Decision-Making (Error)}
\begin{table*}[!h]
\centering
\centering\begingroup\fontsize{8}{10}\selectfont
\renewcommand{\metricsguide}[0]{\cell{r}{Overall Performance\\Overall Gain\\Group Gains\\Max Disparity\\Rat. Violations\\Imputation Risk\\Options Pruned\\Data Use}}
\resizebox{\linewidth}{!}{
\begin{tabular}{lrlllllll}
\multicolumn{2}{c}{ } & \multicolumn{2}{c}{\textsc{Static}} & \multicolumn{2}{c}{\textsc{Imputed}} & \multicolumn{3}{c}{\textsc{Participatory}} \\
\cmidrule(l{3pt}r{3pt}){3-4} \cmidrule(l{3pt}r{3pt}){5-6} \cmidrule(l{3pt}r{3pt}){7-9}
Dataset & Metrics & \OneHot{} & \IntHot{} & \OneHotKNNImpute{} & \IntHotKNNImpute{} & \MinimalSys{} & \FlatSys{} & \SeqSys{}\\
\midrule
\apnea & \metricsguide{} & \cell{r}{26.3\%\\1.5\%\\-0.8\% -- 4.2\%\\5.0\%\\\badvalue{1}\\-1.2\%\\0/6\\100.0\%} & \cell{r}{26.0\%\\1.8\%\\0.4\% -- 3.8\%\\3.4\%\\0\\-1.2\%\\0/6\\100.0\%} & \cell{r}{25.9\%\\1.9\%\\-0.8\% -- 4.2\%\\5.0\%\\\badvalue{1}\\\\0/12\\0.0\%} & \cell{r}{27.4\%\\0.4\%\\0.4\% -- 3.8\%\\3.4\%\\0\\\\0/12\\0.0\%} & \cell{r}{26.3\%\\1.5\%\\0.0\% -- 4.2\%\\4.2\%\\0\\\\2/7\\66.7\%} & \cell{r}{\cellcolor{fgood}\textbf{12.2\%}\\\cellcolor{fgood}\textbf{15.6\%}\\5.3\% -- 22.2\%\\16.9\%\\0\\\\1/12\\100.0\%} & \cell{r}{\cellcolor{fgood}\textbf{12.2\%}\\\cellcolor{fgood}\textbf{15.6\%}\\5.3\% -- 22.2\%\\16.9\%\\0\\\\2/12\\91.7\%}\\
\midrule

\cshockReicu & \metricsguide{} & \cell{r}{18.6\%\\-0.2\%\\-3.5\% -- 1.4\%\\4.9\%\\\badvalue{2}\\-3.5\%\\0/8\\100.0\%} & \cell{r}{17.8\%\\0.6\%\\-2.2\% -- 3.0\%\\5.3\%\\\badvalue{2}\\-2.2\%\\0/8\\100.0\%} & \cell{r}{18.2\%\\0.2\%\\-3.5\% -- 1.4\%\\4.9\%\\\badvalue{2}\\\\0/27\\0.0\%} & \cell{r}{18.6\%\\-0.2\%\\-2.2\% -- 3.0\%\\5.3\%\\\badvalue{2}\\\\0/27\\0.0\%} & \cell{r}{18.4\%\\0.0\%\\0.0\% -- 0.0\%\\0.0\%\\0\\\\8/9\\0.0\%} & \cell{r}{\cellcolor{fgood}\textbf{5.7\%}\\\cellcolor{fgood}\textbf{12.7\%}\\6.0\% -- 14.9\%\\8.9\%\\0\\\\11/27\\100.0\%} & \cell{r}{6.0\%\\12.4\%\\6.0\% -- 14.9\%\\8.9\%\\0\\\\8/27\\91.7\%}\\
\midrule

\cshockRmimic & \metricsguide{} & \cell{r}{19.9\%\\-0.3\%\\-1.1\% -- 1.3\%\\2.4\%\\\badvalue{5}\\-1.1\%\\0/8\\100.0\%} & \cell{r}{20.1\%\\-0.5\%\\-1.3\% -- 0.5\%\\1.7\%\\\badvalue{6}\\-1.3\%\\0/8\\100.0\%} & \cell{r}{19.9\%\\-0.3\%\\-1.1\% -- 1.3\%\\2.4\%\\\badvalue{5}\\\\0/27\\0.0\%} & \cell{r}{20.2\%\\-0.6\%\\-1.3\% -- 0.5\%\\1.7\%\\\badvalue{6}\\\\0/27\\0.0\%} & \cell{r}{19.6\%\\0.0\%\\0.0\% -- 0.0\%\\0.0\%\\0\\\\8/9\\0.0\%} & \cell{r}{11.5\%\\8.1\%\\1.0\% -- 14.9\%\\13.8\%\\0\\\\6/27\\100.0\%} & \cell{r}{\cellcolor{fgood}\textbf{11.4\%}\\\cellcolor{fgood}\textbf{8.1\%}\\1.0\% -- 14.9\%\\13.8\%\\0\\\\5/27\\87.5\%}\\
\midrule

\seercrcagesexcalifornia & \metricsguide{} & \cell{r}{37.2\%\\-0.2\%\\-0.7\% -- 0.1\%\\0.7\%\\\badvalue{4}\\-0.7\%\\0/6\\100.0\%} & \cell{r}{37.0\%\\0.0\%\\-0.3\% -- 0.2\%\\0.5\%\\\badvalue{1}\\-0.3\%\\0/6\\100.0\%} & \cell{r}{37.2\%\\-0.2\%\\-0.7\% -- 0.1\%\\0.7\%\\\badvalue{4}\\\\0/12\\0.0\%} & \cell{r}{37.0\%\\-0.0\%\\-0.3\% -- 0.2\%\\0.5\%\\\badvalue{1}\\\\0/12\\0.0\%} & \cell{r}{37.0\%\\0.0\%\\0.0\% -- 0.0\%\\0.0\%\\0\\\\6/7\\0.0\%} & \cell{r}{\cellcolor{fgood}\textbf{35.9\%}\\\cellcolor{fgood}\textbf{1.0\%}\\0.1\% -- 3.2\%\\3.1\%\\0\\\\3/12\\100.0\%} & \cell{r}{35.9\%\\1.0\%\\0.1\% -- 3.2\%\\3.1\%\\0\\\\5/12\\75.0\%}\\
\midrule

\seerrespacmcollapsedagesexall & \metricsguide{} & \cell{r}{20.0\%\\0.1\%\\-0.3\% -- 0.2\%\\0.6\%\\\badvalue{1}\\-0.3\%\\0/6\\100.0\%} & \cell{r}{20.2\%\\-0.1\%\\-0.5\% -- 0.0\%\\0.5\%\\\badvalue{4}\\-0.5\%\\0/6\\100.0\%} & \cell{r}{20.0\%\\0.1\%\\-0.3\% -- 0.2\%\\0.6\%\\\badvalue{1}\\\\0/12\\0.0\%} & \cell{r}{20.3\%\\-0.2\%\\-0.5\% -- 0.0\%\\0.5\%\\\badvalue{4}\\\\0/12\\0.0\%} & \cell{r}{20.0\%\\0.1\%\\0.0\% -- 0.2\%\\0.2\%\\0\\\\3/7\\50.0\%} & \cell{r}{\cellcolor{fgood}\textbf{19.3\%}\\\cellcolor{fgood}\textbf{0.8\%}\\0.0\% -- 2.3\%\\2.3\%\\0\\\\1/12\\100.0\%} & \cell{r}{19.3\%\\0.7\%\\0.0\% -- 2.2\%\\2.1\%\\0\\\\3/12\\83.3\%}\\
\midrule

\saps & \metricsguide{} & \cell{r}{14.1\%\\0.9\%\\-0.8\% -- 3.4\%\\4.2\%\\\badvalue{1}\\-0.8\%\\0/4\\100.0\%} & \cell{r}{15.0\%\\-0.0\%\\-0.5\% -- 0.3\%\\0.8\%\\\badvalue{1}\\-0.7\%\\0/4\\100.0\%} & \cell{r}{14.1\%\\0.9\%\\-0.8\% -- 3.4\%\\4.2\%\\\badvalue{1}\\\\0/9\\0.0\%} & \cell{r}{15.7\%\\-0.7\%\\-0.5\% -- 0.3\%\\0.8\%\\\badvalue{1}\\\\0/9\\0.0\%} & \cell{r}{13.9\%\\1.1\%\\0.0\% -- 3.4\%\\3.4\%\\0\\\\2/5\\50.0\%} & \cell{r}{\cellcolor{fgood}\textbf{9.8\%}\\\cellcolor{fgood}\textbf{5.2\%}\\0.0\% -- 16.4\%\\16.4\%\\0\\\\1/9\\75.0\%} & \cell{r}{\cellcolor{fgood}\textbf{9.8\%}\\\cellcolor{fgood}\textbf{5.2\%}\\0.0\% -- 16.4\%\\16.4\%\\0\\\\1/9\\87.5\%}\\

\bottomrule
\end{tabular}}
\endgroup{}

\caption{Overview of performance, data use, and consent for all personalized models and systems on all datasets as measured by \textbf{test error}. We show the performance of models and systems built using \textbf{random forests}.}
\end{table*}

\clearpage
\subsection{Random Forests for Ranking (AUC)}
\begin{table*}[!h]
\centering
\centering\begingroup\fontsize{8}{10}\selectfont
\renewcommand{\metricsguide}[0]{\cell{r}{Overall Performance\\Overall Gain\\Group Gains\\Max Disparity\\Rat. Violations\\Imputation Risk\\Options Pruned\\Data Use}}
\resizebox{\linewidth}{!}{
\begin{tabular}{lrlllllll}
\multicolumn{2}{c}{ } & \multicolumn{2}{c}{\textsc{Static}} & \multicolumn{2}{c}{\textsc{Imputed}} & \multicolumn{3}{c}{\textsc{Participatory}} \\
\cmidrule(l{3pt}r{3pt}){3-4} \cmidrule(l{3pt}r{3pt}){5-6} \cmidrule(l{3pt}r{3pt}){7-9}
Dataset & Metrics & \OneHot{} & \IntHot{} & \OneHotKNNImpute{} & \IntHotKNNImpute{} & \MinimalSys{} & \FlatSys{} & \SeqSys{}\\
\midrule
\apnea & \metricsguide{} & \cell{r}{0.825\\0.008\\-0.004 -- 0.009\\0.012\\\badvalue{2}\\-0.004\\0/6\\100.0\%} & \cell{r}{0.824\\0.006\\-0.005 -- 0.012\\0.017\\\badvalue{3}\\-0.005\\0/6\\100.0\%} & \cell{r}{0.822\\0.004\\-0.004 -- 0.009\\0.012\\\badvalue{2}\\\\0/12\\0.0\%} & \cell{r}{0.806\\-0.012\\-0.005 -- 0.012\\0.017\\\badvalue{3}\\\\0/12\\0.0\%} & \cell{r}{0.823\\0.005\\0.000 -- 0.009\\0.009\\0\\\\3/7\\50.0\%} & \cell{r}{\cellcolor{fgood}\textbf{0.944}\\\cellcolor{fgood}\textbf{0.126}\\0.058 -- 0.157\\0.098\\0\\\\2/12\\100.0\%} & \cell{r}{0.942\\0.124\\0.058 -- 0.157\\0.098\\0\\\\4/12\\75.0\%}\\
\midrule

\cshockReicu & \metricsguide{} & \cell{r}{0.896\\0.003\\-0.008 -- 0.011\\0.020\\\badvalue{3}\\-0.008\\0/8\\100.0\%} & \cell{r}{0.896\\0.003\\-0.005 -- 0.011\\0.016\\\badvalue{4}\\-0.005\\0/8\\100.0\%} & \cell{r}{0.897\\0.004\\-0.008 -- 0.011\\0.020\\\badvalue{3}\\\\0/27\\0.0\%} & \cell{r}{0.886\\-0.007\\-0.005 -- 0.011\\0.016\\\badvalue{4}\\\\0/27\\0.0\%} & \cell{r}{0.894\\0.001\\0.000 -- 0.004\\0.004\\0\\\\7/9\\12.5\%} & \cell{r}{\cellcolor{fgood}\textbf{0.987}\\\cellcolor{fgood}\textbf{0.094}\\0.010 -- 0.132\\0.122\\0\\\\10/27\\100.0\%} & \cell{r}{0.987\\0.094\\0.010 -- 0.130\\0.120\\0\\\\10/27\\87.5\%}\\
\midrule

\cshockRmimic & \metricsguide{} & \cell{r}{0.884\\0.000\\-0.005 -- 0.006\\0.011\\\badvalue{3}\\-0.005\\0/8\\100.0\%} & \cell{r}{0.883\\-0.001\\-0.006 -- 0.013\\0.019\\\badvalue{7}\\-0.006\\0/8\\100.0\%} & \cell{r}{0.884\\0.001\\-0.005 -- 0.006\\0.011\\\badvalue{3}\\\\0/27\\0.0\%} & \cell{r}{0.881\\-0.002\\-0.006 -- 0.013\\0.019\\\badvalue{7}\\\\0/27\\0.0\%} & \cell{r}{0.885\\0.001\\0.000 -- 0.006\\0.006\\0\\\\5/9\\37.5\%} & \cell{r}{\cellcolor{fgood}\textbf{0.955}\\\cellcolor{fgood}\textbf{0.071}\\0.016 -- 0.108\\0.092\\0\\\\6/27\\100.0\%} & \cell{r}{0.954\\0.071\\0.016 -- 0.107\\0.090\\0\\\\6/27\\83.3\%}\\
\midrule

\seercrcagesexcalifornia & \metricsguide{} & \cell{r}{0.684\\0.002\\-0.002 -- 0.004\\0.006\\0\\-0.002\\0/6\\100.0\%} & \cell{r}{0.682\\0.000\\-0.004 -- 0.002\\0.007\\0\\-0.004\\0/6\\100.0\%} & \cell{r}{0.681\\-0.001\\-0.002 -- 0.004\\0.006\\0\\\\0/12\\0.0\%} & \cell{r}{0.680\\-0.002\\-0.004 -- 0.002\\0.007\\0\\\\0/12\\0.0\%} & \cell{r}{0.683\\0.001\\0.000 -- 0.004\\0.004\\0\\\\3/7\\50.0\%} & \cell{r}{\cellcolor{fgood}\textbf{0.696}\\\cellcolor{fgood}\textbf{0.014}\\0.004 -- 0.035\\0.030\\0\\\\2/12\\100.0\%} & \cell{r}{0.696\\0.014\\0.004 -- 0.031\\0.026\\0\\\\5/12\\75.0\%}\\
\midrule

\seerrespacmcollapsedagesexall & \metricsguide{} & \cell{r}{0.849\\0.002\\-0.001 -- 0.003\\0.004\\\badvalue{1}\\-0.001\\0/6\\100.0\%} & \cell{r}{0.849\\0.001\\-0.001 -- 0.002\\0.003\\\badvalue{1}\\-0.001\\0/6\\100.0\%} & \cell{r}{0.848\\0.001\\-0.001 -- 0.003\\0.004\\\badvalue{1}\\\\0/12\\0.0\%} & \cell{r}{0.849\\0.001\\-0.001 -- 0.002\\0.003\\\badvalue{1}\\\\0/12\\0.0\%} & \cell{r}{0.848\\0.000\\0.000 -- 0.003\\0.003\\0\\\\2/7\\66.7\%} & \cell{r}{\cellcolor{fgood}\textbf{0.856}\\\cellcolor{fgood}\textbf{0.008}\\0.002 -- 0.020\\0.018\\0\\\\1/12\\100.0\%} & \cell{r}{\cellcolor{fgood}\textbf{0.856}\\\cellcolor{fgood}\textbf{0.008}\\0.002 -- 0.020\\0.018\\0\\\\2/12\\91.7\%}\\
\midrule

\saps & \metricsguide{} & \cell{r}{0.921\\0.003\\-0.002 -- 0.010\\0.012\\\badvalue{2}\\-0.002\\0/4\\100.0\%} & \cell{r}{0.922\\0.004\\-0.002 -- 0.013\\0.015\\\badvalue{2}\\-0.002\\0/4\\100.0\%} & \cell{r}{0.922\\0.003\\-0.002 -- 0.010\\0.012\\\badvalue{2}\\\\0/9\\0.0\%} & \cell{r}{0.906\\-0.012\\-0.002 -- 0.013\\0.015\\\badvalue{2}\\\\0/9\\0.0\%} & \cell{r}{0.921\\0.002\\0.000 -- 0.010\\0.010\\0\\\\2/5\\50.0\%} & \cell{r}{\cellcolor{fgood}\textbf{0.966}\\\cellcolor{fgood}\textbf{0.048}\\0.009 -- 0.109\\0.100\\0\\\\2/9\\100.0\%} & \cell{r}{\cellcolor{fgood}\textbf{0.966}\\\cellcolor{fgood}\textbf{0.048}\\0.009 -- 0.109\\0.100\\0\\\\2/9\\87.5\%}\\

\bottomrule
\end{tabular}}
\endgroup{}

\caption{Overview of performance, data use, and consent for all personalized models and systems on all datasets as measured by \textbf{test auc}. We show the performance of models and systems built using \textbf{random forests}.}
\end{table*}

\end{document}